\documentclass[10pt,journal,compsoc]{IEEEtran}
\ifCLASSOPTIONcompsoc
  \usepackage[nocompress]{cite}
\else
  \usepackage{cite}
\fi

\ifCLASSINFOpdf
\usepackage[pdftex]{graphicx}
\else
\fi
\usepackage{amsmath}

\usepackage{bm}
\usepackage[dvipsnames]{xcolor}
\usepackage{newtxmath}
\usepackage{multirow}
\usepackage{multicol}
\usepackage{adjustbox}
\usepackage{booktabs}
\usepackage{makecell}
\usepackage{float}
\usepackage{caption}
\usepackage{latexsym}
\usepackage{subcaption}
\usepackage{makecell}
\usepackage{pifont} 
\newcommand{\cmark}{\ding{51}} 
\newcommand{\xmark}{\ding{55}}
\usepackage{fixltx2e}
\newcommand{\etal}{\textit{et al.}}

\usepackage[pagebackref=true,breaklinks=true,colorlinks,bookmarks=false]{hyperref}

\begin{document}
%\title{RelTR: End-to-End Scene Graph Generation with Transformers}
%

\title{RelTR: Relation Transformer for Scene Graph Generation}

\author{Yuren Cong, Michael Ying Yang, and Bodo Rosenhahn% <-this % stops a space
\IEEEcompsocitemizethanks{\IEEEcompsocthanksitem Yuren Cong and Bodo Rosenhahn are with Institute of Information Processing, Leibniz University Hannover, Germany.
E-mail: \url{cong@tnt.uni-hannover.de}, \url{rosenhahn@tnt.uni-hannover.de}.
\IEEEcompsocthanksitem Micheal Ying Yang (Corresponding Author) is with Scene Understanding Group, Faculty of Geo-Information Science and Earth Observation (ITC), University of Twente, The Netherlands. Email:
\url{michael.yang@utwente.nl}.}% <-this % stops an unwanted space
}

\IEEEtitleabstractindextext{%
\begin{abstract}
Different objects in the same scene are more or less related to each other, but only a limited number of these relationships are noteworthy.
Inspired by Detection Transformer, which excels in object detection, we view scene graph generation as a set prediction problem.
In this paper, we propose an end-to-end scene graph generation model Relation Transformer (RelTR), which has an encoder-decoder architecture.
The encoder reasons about the visual feature context while the decoder infers a fixed-size set of triplets subject-predicate-object using different types of attention mechanisms with coupled subject and object queries. 
We design a set prediction loss performing the matching between the ground truth and predicted triplets for the end-to-end training.
In contrast to most existing scene graph generation methods, RelTR is a one-stage method that predicts sparse scene graphs directly only using visual appearance without combining entities and labeling all possible predicates. 
Extensive experiments on the Visual Genome, Open Images V6, and VRD datasets demonstrate the superior performance and fast inference of our model.
\end{abstract}

\begin{IEEEkeywords}
Scene Understanding, Scene Graph Generation, One-Stage, Visual Relationship Detection  
\end{IEEEkeywords}
}

\maketitle

\IEEEdisplaynontitleabstractindextext

% \fi
\IEEEpeerreviewmaketitle

\IEEEraisesectionheading{\section{Introduction}\label{sec:introduction}}

\IEEEPARstart{I}{n} scene understanding, a scene graph is a graph structure whose nodes are the entities that appear in the image and whose edges represent the relationships between entities \cite{johnson2015image}.
Scene graph generation (SGG) is a semantic understanding task that goes beyond object detection and is closely linked to visual relationship detection \cite{lu2016visual}. 
At present, scene graphs have shown their potential in different vision-language tasks such as image retrieval \cite{johnson2015image}, image captioning \cite{Nguyen_2021_ICCV,gao2018image}, visual question answering (VQA) \cite{johnson2017inferring} and image generation \cite{ashual2019specifying, johnson2018image}. 
The task of scene graph generation has also received sustained attention in the computer vision community. 
%%%%%%%%%%%%%%%%%%
\begin{figure}[!t]
\centering
\includegraphics[width=0.99\linewidth]{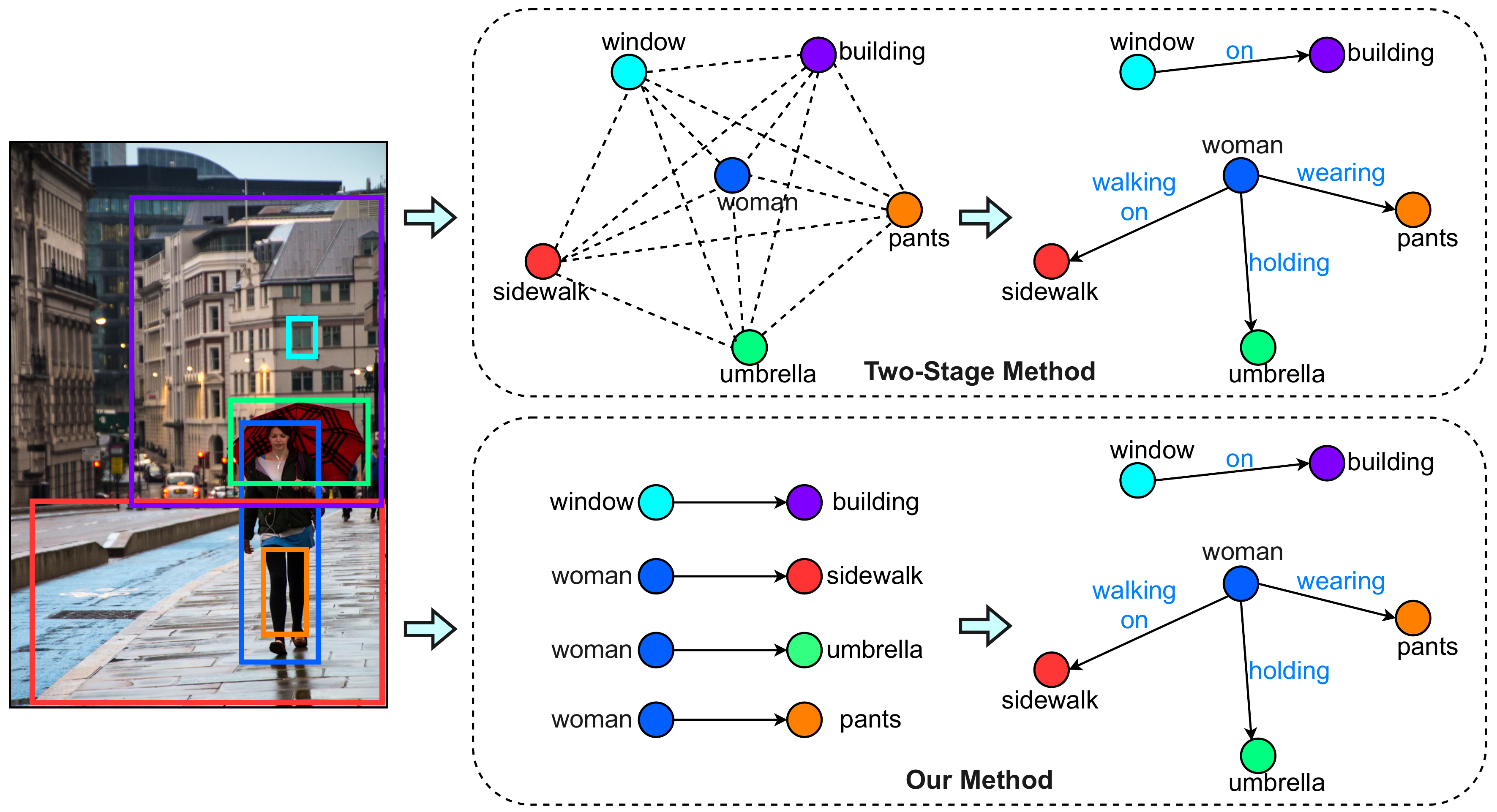}
\caption{Different from most existing two-stage methods that label the dense relationships between all entity proposals, our one-stage approach can predict the pair proposals directly and generate a sparse scene graph with only visual appearance.}
\label{fig:teaser}
\end{figure}
%%%%%%%%%%%%%%%%%%
Most existing methods for generating scene graphs employ an object detector (e.g. FasterRCNN \cite{ren2016faster}) and use some specific neural networks to infer the relationships. 
The object detector generates proposals in the first stage,  and the relationship classifier labels the edges between the object proposals for the second stage. 
Although these two-stage approaches have made incredible progress, they still suffer from the drawback that these models require a large number of trained parameters. 
% Moreover, if $n$ object proposals are given, the relationship inference network runs the risk of learning based on erroneous features provided by the detection backbone and has to predict $\mathcal{O}(n^2)$ relationships, which leads to slow inference (see Fig.~\ref{fig:teaser}).
If $n$ object proposals are given, the relationship inference network runs the risk of learning based on erroneous features provided by the detection backbone and has to predict $\mathcal{O}(n^2)$ relationships (see Fig.~\ref{fig:teaser}).
This manipulation may lead to the selection of triplets based on the confident scores of object proposals rather than interest in relationships.
%%%%%%%
Many previous works  \cite{zellers2018neural, chen2019knowledge, yu2017visual,zareian2020bridging, gu2019scene} have integrated semantic knowledge to improve their performance. 
However, these models face significant biases in relationship inference conditional on subject and object categories. 
They prefer to predict the predicates that are popular between particular subjects and objects, rather than those based on visual appearance.

Recently, the one-stage models have emerged in the field of object detection \cite{law2018cornernet,tian2019fcos, zhou2019objects, sun2021makes}. 
They are attractive for the fast speed, low costs, and simplicity. These are also the properties that are urgently needed for the scene graph generation models. 
Detection Transformer (DETR)~\cite{carion2020end} views object detection as an end-to-end set prediction task and proposes a set-based loss via bipartite matching. 
This strategy can be extended to scene graph generation: based on a set of learned subject and object queries, a fixed number of triplets \texttt{$<$subject-predicate-object$>$} could be predicted by reasoning about the global image context and co-occurrences of entities. 
However, it is challenging to implement such an intuitive idea. 
The model needs to predict both the location and the category of the subject and object, and also consider their semantic connection. Furthermore, the direct bipartite matching is not competent to assign ground truth information to relationship predictions. This paper aims to address these challenges. 

We propose a novel end-to-end framework for scene graph generation, named \textbf{Relation Transformer (RelTR)}. %~\footnote{The source code is  publicly available at \url{https://github.com/yrcong/RelTR}.}.
As shown in Fig.~\ref{fig:teaser}, RelTR can detect the triplet proposals with only visual appearance and predict subjects, objects, and their predicates concurrently. 
We evaluate RelTR on Visual Genome \cite{krishna2017visual} and large-scale Open Images V6 \cite{kuznetsova2020open}. 
The \textbf{main contributions} of this work are summarized as follows: 
\begin{itemize}
    \item In contrast to most existing advanced approaches that classify the dense relationships between all entity proposals from the object detection backbone, our one-stage method can generate a sparse scene graph by decoding the visual appearance with the subject and object queries learned from the data.
    \item RelTR generates scene graphs based on visual appearance only, which has fewer parameters and faster inference compared to other SGG models while achieving state-of-the-art performance. 
    \item  A set prediction loss is designed to perform the matching between the ground truth and predicted triplets with an IoU-based assignment strategy. 
    \item  With the decoupled entity attention, the triplet decoder of RelTR can improve the localization and classification of subjects and objects with the entity detection results from the entity decoder. 
    \item  Through comprehensive experiments, we explore which components are critical for the performance and analyze the working mechanism of learned subject and object queries.
    \item RelTR can be simply implemented. The source code and pretrained model are publicly available at \url{https://github.com/yrcong/RelTR}.
\end{itemize}

The remainder of the paper is structured as follows.
In Section~\ref{sec:related}, we review related work in scene graph generation. 
Section~\ref{sec:method} presents our proposed method. 
Experimental results of the proposed framework are discussed in Section~\ref{sec:exp}.
Section~\ref{sec:con} concludes this paper.

%%%%%%%%%%%%%%%%%%
\begin{figure*}[http]
\centering
\includegraphics[width=1\linewidth]{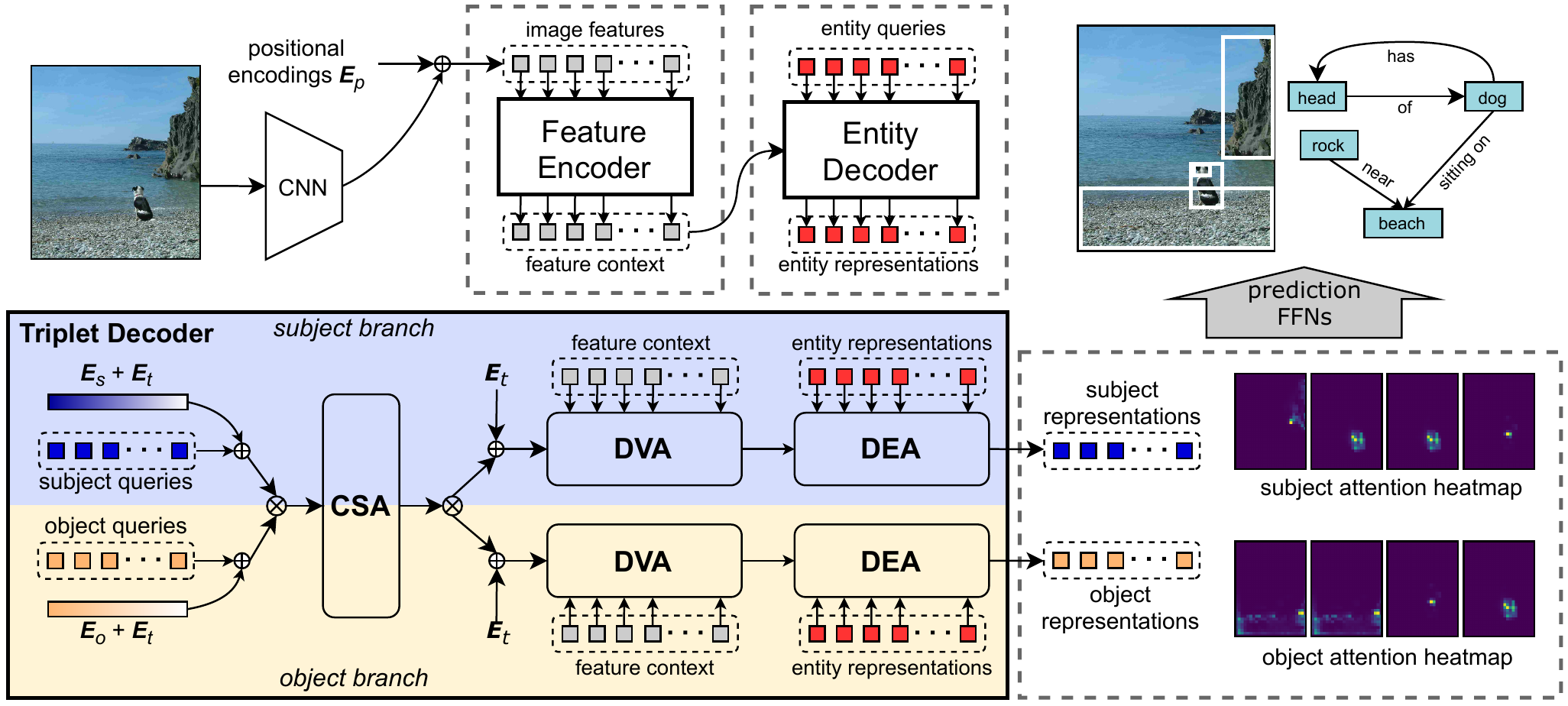}
\caption{Given a set of learned subject and object queries coupled by subject and object encodings, RelTR captures the dependencies between relationships and reasons about the feature context and entity representations, respectively the output of the feature encoder and entity decoder, to directly compute a set of subject and object representations. 
A pair of subject and object representations with attention heat maps is decoded into a triplet \texttt{$<$subject-predicate-object$>$} by feed forward networks (FFNs). 
\textbf{CSA}, \textbf{DVA} and \textbf{DEA} stand for Coupled Self-Attention, Decoupled Visual Attention and Decoupled Entity Attention.  
$\bm{E_p}$, $\bm{E_t}$, $\bm{E_s}$ and $\bm{E_o}$ are  the positional, triplet, subject and object encodings respectively. 
$\oplus$ indicates element-wise addition, while $\otimes$ indicates concatenation or split.} 
\label{fig:framework}
\end{figure*}  
%%%%%%%%%%%%%%%%%%

\section{Related Work}
\label{sec:related}
\subsection{Scene Graph Generation}
Scene graphs have been proposed in \cite{johnson2015image} for the task of image retrieval and attract increasing attention in computer vision and natural language processing communities for different scene understanding tasks such as image captioning \cite{yang2019auto, gu2019unpaired, lee2019learning}, VQA \cite{shi2019explainable, lee2019visual} and image synthesis \cite{li2019pastegan, talavera2019layout}. 
The main purpose of scene graph generation (SGG) is to detect the relationships between objects in the scene. 
Many earlier works were limited to identifying specific types of relationships such as spatial relationships between entities \cite{galleguillos2008object,gould2008multi}.
The universal visual relationship detection is introduced in \cite{lu2016visual}.
Their inference framework, which detects entities in an image first and then determines dense relationships, was widely adopted in subsequent works, including their evaluation settings and metrics as well.

Now many models \cite{cong2020nodis, wang2019exploring, shi2021simple,wang2021topic,lu2021context,tang2019learning, chen2019counterfactual,chiou2021recovering} are available to generate scene graphs from different perspectives, and some works even extend the scene graph generation task from images to videos \cite{ji2020action,cong2021spatial, teng2021target, lu2021multi}. 
To solve the problem of class imbalance, several unbiased scene graph generation methods are recently proposed \cite{suhail2021energy,yan2020pcpl,guo2021general,desai2021learning}.
Two-stage methods following \cite{lu2016visual} are currently dominating scene graph generation: 
several works \cite{zellers2018neural, xu2017scene,li2017scene, cong2020nodis} use residual neural networks with the global context to improve the quality of the generated scene graphs. 
Xu \etal~\cite{xu2017scene} use standard RNNs to iteratively improve the relationship prediction via message passing while MotifNet \cite{zellers2018neural} stacks LSTMs to reason about the local and global context.
Graph-based models \cite{yang2018graph,li2018factorizable,li2021bipartite,chen2019knowledge,lin2020gps} perform message passing and demonstrate good results.
Factorizable Net ~\cite{li2018factorizable} decomposes and combines the graphs to infer the relationships.
The attention mechanism is integrated into different types of graph-based models such as Graph R-CNN~\cite{yang2018graph},
GPI~\cite{herzig2018mapping} and ARN~\cite{qi2019attentive}.
With the rise of Transformer \cite{vaswani2017attention}, there are several attempts using Transformer to detect visual relationships and generate scene graphs in very recent works \cite{dhingra2021bgt,koner2020relation,lu2021context}.
RelTransformer \cite{chen2022reltransformer} tackles the compositionality in visual relationship recognition  with an effective message-passing flow.
To improve performance, many works are no longer limited to using only visual appearance. 
Semantic knowledge can be utilized as an additional feature to infer scene graphs \cite{lu2016visual, zellers2018neural, gkanatsios2019attention, cui2018context, yu2017visual, yu2020cogtree}. 
Furthermore, statistic priors and knowledge graphs have been introduced in \cite{zhang2019graphical,dai2017detecting,yu2017visual}.

Compared to the boom of two-stage approaches, one-stage approaches are still in their infancy and have the advantage of being simple, fast, and easy to train. 
%To the best of our knowledge, 
FCSGG \cite{liu2021fully} is a one-stage scene graph generation framework that encodes objects as box center points and relationships as 2D vector fields.
While FCSGG model being lightweight and fast speed, it has a significant performance gap compared to other two-stage methods. 
To fill this gap, we propose Transformer-based RelTR using only visual appearance in this work with fewer parameters, faster inference speed, and higher accuracy. 
Recently, SGTR \cite{li2022sgtr} also introduces an end-to-end framework predicting entity and predicate proposals independently. 
A graph assembling module is designed to connect the entity and predicates.
In contrast, our RelTR directly predicts triplet proposals and achieves higher recall scores. 
Distinct from the other two-stage Transformer-based approaches~\cite{dhingra2021bgt,koner2020relation,lu2021context} that utilize the attention mechanism to capture the context of the entity proposals from an object detector, RelTR can decode the global feature maps directly with the subject and object queries learned from the data to generate a sparse scene graph.

\subsection{Transformer and Set Prediction}
The original Transformer architecture was proposed in~\cite{vaswani2017attention} for sequence transduction. 
Its encoder-decoder configuration and attention mechanism is also used to solve various vision tasks in different ways,~e.g. object detection \cite{carion2020end}, image pre-training \cite{yuan2022rlip}, human-object interaction (HOI) detection \cite{kim2021hotr}, and dynamic scene graph generation \cite{cong2021spatial}.

DETR \cite{carion2020end} is a seminal work based on Transformer architecture for object detection in recent years. 
It views detection as a set prediction problem. 
In the end-to-end training, with the object queries, DETR predicts a fixed-size set of object proposals and performs a bipartite matching between proposals and ground truth objects for the loss function. 
This  concept of query-based set prediction quickly gains popularity in the computer vision community. 
Many tasks can be reformulated as set prediction problems,~e.g.~instance segmentation \cite{wang2021end}, image captioning \cite{liu2021cptr} and multiple-object tracking \cite{zeng2021motr}.
Some works \cite{zhu2020deformable, yao2021efficient} attempt to further improve object detection based on DETR.

% dong2021visual
HOI detection localizes and recognizes the relationships between humans and objects, whose result is a sub-graph of the scene graph. 
Several HOI detection frameworks \cite{kim2021hotr, zou2021end} have been developed that use holistic triplet queries to directly infer a set of interactions. 
However, such a concept is difficult to generalize to the more complex task of scene graph generation.
On large-scale datasets, such as Visual Genome \cite{krishna2017visual} and Open Images \cite{kuznetsova2020open},  localization and classification of subjects and objects using only triplet queries may likely result in low accuracy.
On the contrary, our proposed RelTR predicts the general relationships using coupled subject and object queries to achieve high accuracy. 

\section{Method}
\label{sec:method}

A scene graph $\mathcal{G}$ consists of entity vertices $\bm{\mathcal{V}}$ and relationship edges $\bm{\mathcal{E}}$. 
Different from previous works that detect a set of entity vertices and label the predicates between the vertices, we propose a one-stage model, Relation Transformer (RelTR), to directly predict a fixed-size set of $<\mathcal{V}_{sub}-\mathcal{E}_{prd}-\mathcal{V}_{obj}>$ for scene graph generation.

\subsection{Preliminaries}
\subsubsection{Transformer} 
We provide a brief review on Transformer and its attention mechanism. Transformer~\cite{vaswani2017attention} has an encoder-decoder structure and consists of stacked attention functions. 
The input of a single-head attention is formed from queries $\bm{Q}$, keys $\bm{K}$ and values $\bm{V}$ while the output is computed as:
%%%%%%%%%%%%%%%%%%
\begin{equation}\label{eq:attention} 
\centering
%\begin{aligned}
  Attention(\bm{Q},\bm{K},\bm{V}) = softmax\left(\frac{\bm{Q}\bm{K}^T}{\sqrt{d_k}}\right) \bm{V},
%\end{aligned}
\end{equation}
%%%%%%%%%%%%%%%%%%
where $d_k$ is the dimension of $\bm{K}$. 
In order to benefit from the information in different representation sub-spaces, multi-head attention is applied in Transformer. 
A complete attention function is a multi-head attention followed by a normalization layer with residual connection and denoted as $Att(.)$ in this paper for simplicity.

\subsubsection{DETR} 
This entity detection framework~\cite{carion2020end} is built upon the standard Transformer encoder-decoder architecture. 
First, a CNN backbone generates a feature map $\bm{Z}_0\in\mathbb{R}^{H \times W \times d}$ for an image. With the self-attention mechanism, the encoder computes a new feature context $\bm{Z}\in\mathbb{R}^{HW \times d}$ using the flatted $\bm{Z}_0$ and fixed positional encodings $\bm{E}_p\in\mathbb{R}^{HW \times d}$. 
The decoder transforms $N_e$ entity queries into the entity representations $\bm{Q}_e\in\mathbb{R}^{N_e \times d}$. 
The entity queries interact with each other to capture the entity context and extract visual features from $\bm{Z}$.% in the decoder. 

For the end-to-end training, a set prediction loss for entity detection is proposed in DETR by assigning the ground truth entities to predictions. 
The ground truth set of size $N_e$ is padded with $\phi$ \texttt{$<$background$>$}, % while $N_e$ is larger than the number of entities in any image. 
and a cost function $c_{m}(\hat{y}, y)$ is applied to compute the matching cost between a prediction $\hat{y}$ and ground truth entity $y=\left\{c,b\right\}$ where $c,b$ indicates the target class and box coordinates respectively. 
Given the cost matrix $\bm{C}_{ent}$, the entity prediction-ground truth assignment is computed with the Hungarian algorithm~\cite{stewart2016end}. The set prediction loss for entity detection can be presented as:
%%%%%%%%%%%%%%%%%%
\begin{equation}
\centering
L_{entity} = \sum_{i=1}^{N_e} \left[ L_{cls} + \vmathbb{1}_{\left\{ c^i\ne \phi \right\} }L_{box} \right],
\label{eq:loss_entity}
\end{equation}
where $L_{cls}$ denotes the cross-entropy loss for label classification and $c^i\ne \phi$ means that \texttt{$<$background$>$} is not assigned to the $i$-th entity prediction. $L_{box}$ consists of  $L_1$ loss and generalized IoU loss \cite{rezatofighi2019generalized} for box regression.
%%%%%%%%%%%%%%%%%%

\subsection{RelTR Model}
As shown in Fig.~\ref{fig:framework}, our one-stage model RelTR has an encoder-decoder architecture, which directly predicts $N_t$ triplets without inferring the possible predicates between all entity pairs. 
It consists of the feature encoder extracting the visual feature context, the entity decoder capturing the entity representations from DETR~\cite{carion2020end}, and the triplet decoder with the subject and object branches.% for the triplet inference.

A triplet decoder layer contains three attention functions,  coupled self-attention (CSA), decoupled visual attention (DVA), and decoupled entity attention (DEA), respectively. 
Given $N_t$ coupled subject and object queries, the triplet decoder layer reasons about the feature context $\bm{Z}$ and entity representations $\bm{Q}_e$ from the entity decoder layer to directly output the information of $N_t$ triplets without inferring the possible predicates between all entity pairs. 

\subsubsection{Subject and Object Queries}
There are two types of learned embeddings, namely subject queries $\bm{Q}_s\in\mathbb{R}^{N_t \times d}$ and object queries $\bm{Q}_o\in\mathbb{R}^{N_t \times d}$, for the subject branch and object branch respectively. 
These $N_t$ pairs of subject and object queries are transformed into $N_t$ pairs of subject and object representations of size $d$. 
However, the subject query and the object query are not actually linked together in a query pair since the attention layers in the triplet decoder are permutation invariant. 
In order to distinguish between different triplets, the learnable triplet encodings $\bm{E}_t\in\mathbb{R}^{N_t \times d}$ are introduced.

\subsubsection{Coupled Self-Attention (CSA)}
Coupled self-attention captures the context between $N_t$ triplets and the dependencies between all subjects and objects. 
Although the triplet encodings $\bm{E}_t$ are already available, we still need subject encodings $\bm{E}_s$ and object encodings $\bm{E}_o$ of the same size as $\bm{E}_t$ to inject the semantic concepts of \texttt{$<$subject$>$} and \texttt{$<$object$>$} in coupled self-attention.
Both $\bm{E}_s$ and $\bm{E}_o$ are randomly initialized and learned in the training.
The subject and object queries are encoded and the output of CSA can be formulated as: 
%%%%%%%%%%%%%%%%%%
\begin{equation}
\centering
\begin{aligned}
& \bm{Q} = \bm{K} = [\bm{Q}_s + \bm{E}_{s}+ \bm{E}_{t}, \bm{Q}_o + \bm{E}_{o}+ \bm{E}_{t}]\\
&[\bm{Q}_s, \bm{Q}_o] = Att_{CSA}(\bm{Q},\bm{K}, [\bm{Q}_s, \bm{Q}_o]),
\end{aligned}
\label{eq:csa}
\end{equation}
where $[,]$ indicates the unordered concatenation operation and the updated embeddings keep the original symbols unchanged for brevity. The output of CSA $[\bm{Q}_s, \bm{Q}_o]$ is decoupled into $\bm{Q}_s$ and $\bm{Q}_o$ which continue to be used for the subject branch and the object branch, respectively.
Coupled self-attention enables the subject queries $\bm{Q}_s$ and object queries $\bm{Q}_o$ aware of each other and provides the preconditions for the following cross-attentions.
%%%%%%%%%%%%%%%%%%
\subsubsection{Decoupled Visual Attention (DVA)} Decoupled visual attention concentrates on extracting visual features from the feature context $\bm{Z}$. \textit{Decoupled} means that the computations of subject and object representations are independent of each other, which is distinct from CSA. 
In the subject branch, $\bm{Q}_s\in\mathbb{R}^{N_t \times d}$ are updated through their interaction with the feature context $\bm{Z}\in\mathbb{R}^{HW \times d}$. 
The feature context combines with fixed position encodings $\bm{E}_p\in\mathbb{R}^{HW \times d}$ again in DVA. 
The updated subject representations containing visual features are presented as:
%%%%%%%%%%%%%%%%%%
\begin{equation}
\centering
\begin{aligned}
& \bm{Q} = \bm{Q}_s + \bm{E}_{t}, \bm{K} = \bm{Z} + \bm{E}_{p} \\
&\bm{Q}_s = Att^{(sub)}_{DVA}(\bm{Q},\bm{K}, \bm{Z}).
\end{aligned}
\label{eq:dva}
\end{equation}
%%%%%%%%%%%%%%%%%%
The same operation is performed in the object branch.
In the multi-head attention operation, $N_t$ attention heat maps $\bm{M}_s\in\mathbb{R}^{N_t \times HW}$ are computed. We also adopt the reshaped heat maps as a spatial feature for predicate classification.

\subsubsection{Decoupled Entity Attention (DEA)} Decoupled entity attention is performed as the bridge between entity detection and triplet detection.
Entity representations $\bm{Q}_e\in\mathbb{R}^{N_e \times d}$ can provide localization and classification information with higher quality due to the fact that they do not have semantic restrictions like those between subject and object representations. 
%The motivation for introducing DEA is the wish for subjects and object embeddings to learn more accurate localization and classification information from entity representations through the attention mechanism. 
The motivation for introducing DEA is expecting subject and object representations to learn more accurate localization and classification information from entity representations through the attention mechanism. 
$\bm{Q}_s$ and $\bm{Q}_o$ are finally updated in a triplet decoder layer as follows:
%%%%%%%%%%%%%%%%%%
\begin{equation}
\centering
\begin{aligned}
&\bm{Q}_s = Att^{(sub)}_{DEA}(\bm{Q}_s+\bm{E}_{t}, \bm{Q}_e, \bm{Q}_e) \\
&\bm{Q}_o = Att^{(obj)}_{DEA}(\bm{Q}_o+\bm{E}_{t}, \bm{Q}_e, \bm{Q}_e),
\end{aligned}
\label{eq:dea}
\end{equation}
%%%%%%%%%%%%%%%%%%
where $Att^{(sub)}_{DEA}$ and $Att^{(obj)}_{DEA}$ are the decoupled entity attention modules in the subject and object branch. The outputs of DEA are processed by a feed-forward network followed by a normalization layer with residual connection. The feed-forward network (FFN) consists of two linear transformation layers with ReLU activation.

%%%%%%%%%%%%%%%%%%
\begin{figure}[http]
\centering
\includegraphics[width=0.99\linewidth]{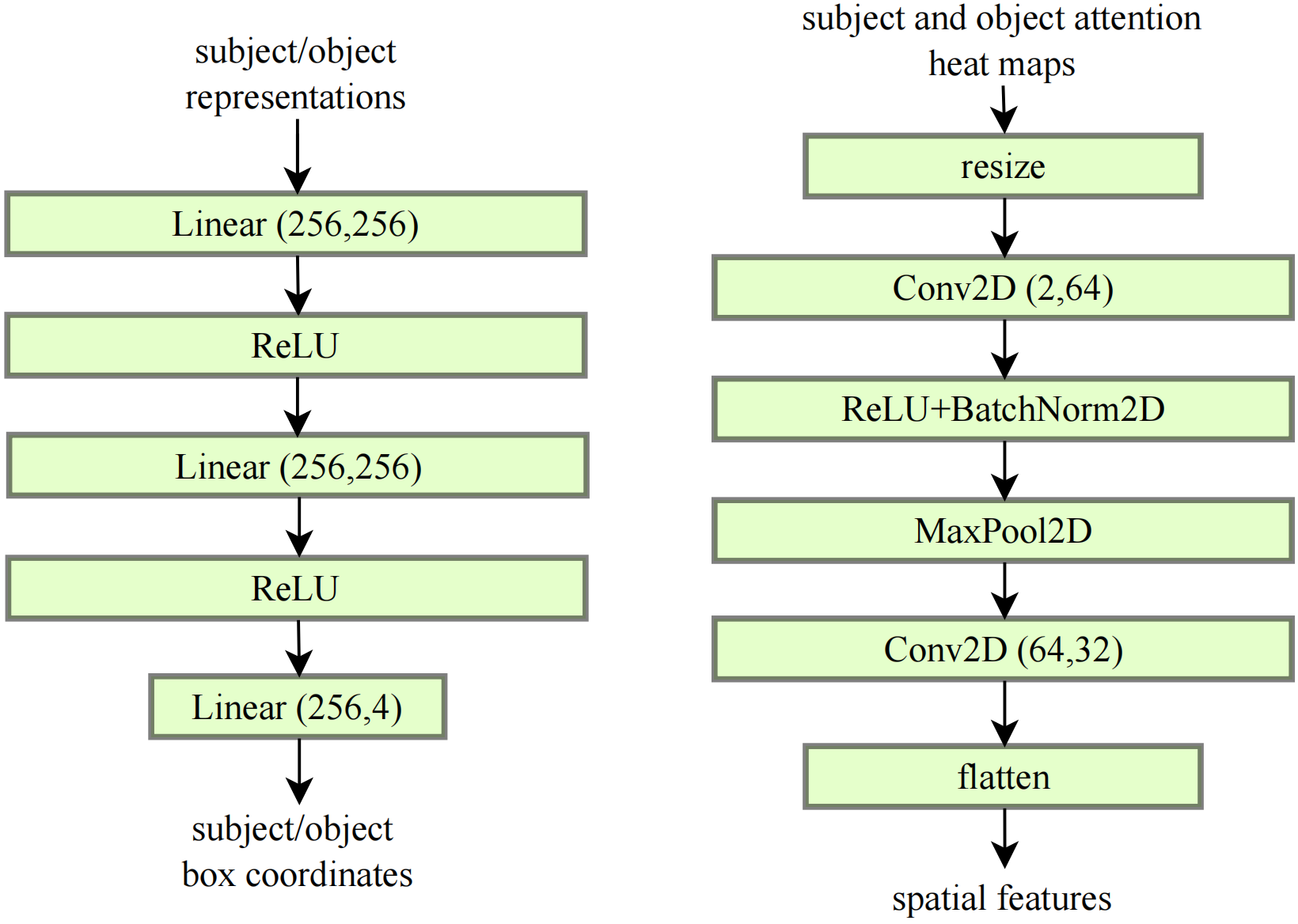}
\caption{\textit{Left}: Architecture of the feed-forward network for subject/object box regression. \textit{Right}: Architecture of the convolutional mask head.} %converting the attention heat maps to spatial feature vectors ().}
\label{fig:ffn}
\end{figure}  
%%%%%%%%%%%%%%%%%%

%%%%%%%%%%%%%%%%%%
\begin{figure*}[http]
\centering
\includegraphics[width=0.99\linewidth]{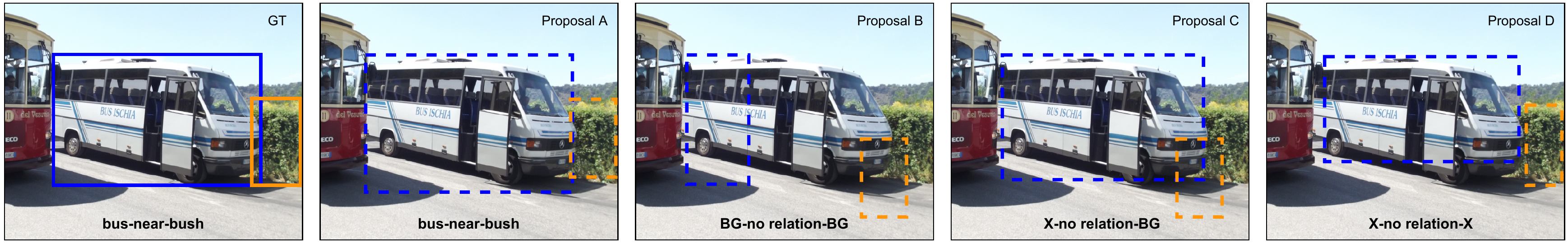}
\caption{The ground truth is assigned to Proposal A while \texttt{$<$background-no relation-background$>$} is assigned to Proposal B. However,  \texttt{$<$background$>$} should not be assigned to the subject of Proposal C and the subject as well as object of Proposal D. \textbf{BG} denotes \texttt{$<$background$>$} while \textbf{X} indicates no assignment.}
%We propose an  IoU-based assignment strategy to prevent this problem.} 
\label{fig:matching}
\end{figure*} 
%%%%%%%%%%%%%%%%%%

\subsubsection{Final Inference}
A complete triplet includes the predicate label and the class labels as well as the bounding box coordinates of the subject and object.
The subject representations $\bm{Q}_s$ and object representations $\bm{Q}_o$ from the last decoder layer are transformed by two linear projection layers into entity class distributions. 
We utilize two independent feed-forward networks with the same structure to predict the height, width, and normalized center coordinates of subject and object boxes. 
The architecture is shown in Fig.~\ref{fig:ffn} (left).
A pair of subject attention heat map $\bm{M}_s$ and object attention heatmap $\bm{M}_o$ from  DVA  modules  in  the  last  decoder  layer  is concatenated and resized $2\times28\times28$. 
%A convolutional mask head converts $N_t$ attention heat maps $\bm{M}_s$ and $\bm{M}_o$ from DVA modules in the last decoder layer to $N_t$ spatial feature vectors. 
The convolutional mask head shown in Fig.~\ref{fig:ffn} (right) converts the attention heat maps to spatial feature vectors $\bm{V}_{spa}$.
The predicate probability $\hat{\bm{p}}_{prd}$ is predicted by a multi-layer perceptron concatenating the corresponding subject representation, object representation, and spatial feature vector, which can be formulated as:
\begin{equation}
\centering
\hat{\bm{p}}_{prd} = \text{softmax}( \text{MLP}([\bm{Q}_s, \bm{Q}_o, \bm{V}_{spa}])).
\label{eq:final_prediction}
\end{equation}
The final predicate labels are determined based on the predicted probabilities.

\subsection{Set Prediction Loss for Triplet Detection}

We design a set prediction loss for triplet detection by extending the entity detection set prediction loss in Eq.~\ref{eq:loss_entity}. 
%We extend the entity detection set prediction loss in Eq.~\ref{eq:loss_entity} to the set prediction loss for triplet detection. 
We present a triplet prediction as $\left \langle \hat{y}_{sub},\hat{c}_{prd},\hat{y}_{obj} \right \rangle$ where $\hat{y}_{sub}=\left\{\hat{c}_{sub},\hat{b}_{sub} \right\}$ and $\hat{y}_{obj}=\left\{\hat{c}_{obj},\hat{b}_{obj} \right\}$ while a ground truth is denoted as $\left \langle y_{sub},c_{prd},y_{obj} \right\rangle$.
The predicted subject, predicate, and object labels are respectively denoted as $\hat{c}_{sub}$, $\hat{c}_{prd}$ and $\hat{c}_{obj}$ while the predicted box coordinates of the subject and object are denoted as $\hat{b}_{sub}$ and $\hat{b}_{obj}$.

When $N_t$ relationships are predicted and $N_t$ is larger than the number of triplets in the
image, the ground truth set of triplets is padded with $\Phi$ \texttt{$<$background-no relation-background$>$}. 
The pair-wise matching cost $c_{tri}$ between a predicted triplet and a ground truth triplet consists of the subject cost $c_{m}(\hat{y}_{sub},y_{sub})$, object cost $c_{m}(\hat{y}_{obj},y_{obj})$ and predicate cost $c_{m}(\hat{c}_{prd},c_{prd})$.
The prediction $\hat{y}=\left\{ \hat{c}, \hat{b} \right\}$ contains the predicted class $\hat{c}$ including the class probabilities $\hat{\bm{p}}$ and the predicted box coordinates $\hat{b}$ while the ground truth $y=\left\{c, b\right\}$ contains the ground truth class $c$ and the ground truth box $b$.
For the predicate, we only have the predicted class $\hat{c}_{prd}$ and ground truth class $c_{prd}$.

The subject/object cost is determined by the predicted entity class probability and the predicted bounding box while the predicate cost is determined only by the predicted predicate class probability.
We define the predicted probability of class $c$ as $\hat{\bm{p}}(c)$.
We adopt the class cost function from~\cite{zhu2020deformable} which can be formulated as:
\begin{equation}
\centering
\begin{aligned}
&c^{+}_{cls}(\hat{c},c) = \alpha\cdot(1-\hat{\bm{p}}(c))^{\gamma}\cdot (-log(\hat{\bm{p}}(c)+\varepsilon)) \\
&c^{-}_{cls}(\hat{c},c) =(1-\alpha)\cdot \hat{\bm{p}}(c)^{\gamma}\cdot (-log(1-\hat{\bm{p}}(c)+\varepsilon)) \\
&c_{cls}(\hat{c},c) = c^{+}_{cls}(\hat{c},c)-c^{-}_{cls}(\hat{c},c),
\end{aligned}
\label{eq:cls_loss}
\end{equation}
where $\alpha$, $\gamma$ and $\varepsilon$ are respectively set to $0.25$, $2$ and $10^{-8}$. 
The box cost for the subject and object is computed using  $L_1$ loss and  generalized IoU loss \cite{rezatofighi2019generalized}:
\begin{equation}
\centering
\begin{aligned}
&c_{box}(\hat{b},b) = 5L_1(\hat{b},b)+2L_{GIoU}(\hat{b},b).
\end{aligned}
\label{eq:box_loss}
\end{equation}
The cost function $c_m$ can be presented as:
\begin{equation}
\centering
\begin{aligned}
&c_m(\hat{y},y) = c_{cls}(\hat{c},c)+ \vmathbb{1}_{\left\{ b \in y \right\}}c_{box}(\hat{b},b), \\
\end{aligned}
\label{eq:c_m}
\end{equation}
where $ b \in y$ denotes that the ground truth includes the box coordinates (only for the subject/object cost).
The cost between a triplet prediction and a ground truth triplet is computed as:
\begin{equation}
\centering
\begin{aligned}
&c_{tri} = c_{m}(\hat{y}_{sub},y_{sub})+c_{m}(\hat{y}_{obj},y_{obj})+c_{m}(\hat{c}_{prd},c_{prd}),  \\
\end{aligned}
\label{eq:cost_triplet}
\end{equation}

%%%%%%%%%%%%%%%%%%
\begin{table*}[t!]
\centering
\begin{adjustbox}{width=1\textwidth}
\begin{tabular}{c|c|c|cccc|cccc|cccc|c|c}
 \hline \hline
 \multicolumn{2}{c|}{\multirow{2}*{Method}}& \multirow{2}*{$AP_{50}$} & \multicolumn{4}{c|}{PredCLS $\uparrow$} & \multicolumn{4}{c|}{SGCLS $\uparrow$} & \multicolumn{4}{c|}{SGDET $\uparrow$} & \multirow{2}*{\#params(M) $\downarrow$} & \multirow{2}*{FPS $\uparrow$} \\
  \multicolumn{2}{c|}{} & &R@20 &R@50 &mR@20 &mR@50 &R@20 &R@50 &mR@20 &mR@50 &R@20 &R@50 &mR@20 &mR@50 & & \\
  \hline 
 \multirow{9}*{\makecell[c]{two-\\stage}}
&\textcolor{blue}{MOTIFS} \cite{zellers2018neural}& 20.0 & 58.5& 65.2& 10.8& 14.0& 32.9 & 35.8 & 6.3 & 7.7 & 21.4 & 27.2 & 4.2 & 5.7 & 240.7 & 6.6\\
&\textcolor{blue}{KERN} \cite{chen2019knowledge}& 20.0& 59.1& 65.8& -& 17.7& 32.2 & 36.7 & - & 9.4 & 22.3 & 27.1 & - & 6.4 & 405.2 & 4.6\\
&\textcolor{blue}{PCPL} \cite{yan2020pcpl}& -& -& 50.8& -& \textbf{35.2}& - & 27.6 & - & 18.6  & - & 14.6 & - & 9.5 & - & -\\
&\textcolor{blue}{GB-Net} \cite{zareian2020bridging}& -& -& 66.6& -& 19.3& - & 38.0 & - & 9.6  & - & 26.4 & - & 6.1 & - & -\\
&\textcolor{blue}{RelDN} \cite{zhang2019graphical}& -& 66.9& 68.4& -& -& 36.1 & 36.8 & - & - & 21.1 & 28.3 & - & - & 615.6 & 4.7\\
&\textcolor{blue}{VCTree-TDE} \cite{tang2020unbiased}& 28.1 & 39.1& 49.9& 17.2& 23.3& 22.8 & 28.8 & 8.9 & 11.8 & 14.3 & 19.6 & 6.3 & 9.3 & 360.8 & 1.2\\
&\textcolor{blue}{GPS-Net} \cite{lin2020gps}& -& \textbf{67.6}& \textbf{69.7}& \textbf{17.4}& 21.3& \textbf{41.8} & 42.3 & 10.0 & 11.8 & 22.3 & 28.9 & 6.9 & 8.7 & - & -\\
&\textcolor{blue}{BGNN} \cite{li2021bipartite}& 29.0& - & 59.2& -& \textbf{30.4}& - & 37.4 & - & \textbf{14.3} & 23.3 & 31.0 & \textbf{7.5} & \textbf{10.7} & 341.9 & 2.3\\
&\textcolor{blue}{BGT-Net} \cite{dhingra2021bgt}& 28.1& 60.9 & 67.1& 16.8 & 20.6 & 41.7 & \textbf{45.9} & \textbf{10.4} & 12.8 & \textbf{25.5} & \textbf{32.8} & 5.7 & 7.8 & - & -\\
&IMP \cite{xu2017scene}& -& 58.5& 65.2& -& 9.8 &31.7 & 34.6 & - & 5.8 & 14.6 & 20.7 & - & 3.8 & \textbf{203.8} & \textbf{10.0}\\
&CISC \cite{wang2019exploring}& -& 42.1& 53.2& -& -& 23.3 & 27.8 & - & - & 7.7 & 11.4 & - & - & - & -\\
&G-RCNN \cite{yang2018graph}& 24.8& -& 54.2& -& -& - & 29.6 & - & - & - & 11.4 & - & - & - & -\\
\hline
 \multirow{3}*{\makecell[c]{one-\\stage}}
&FCSGG \cite{liu2021fully}& 28.5& 33.4& 41.0& 4.9& 6.3& 19.0 & 23.5 & 3.7 & 3.7 & 16.1 & 21.3 & 2.7 & 3.6 & 87.1  & 8.4\\
&SGTR \cite{li2022sgtr}& - & - & - &- &-& - &- & - & - & - & 20.6 & - & \textbf{\textit{15.8}} & 166.5  & 3.4\\
&RelTR (ours) & 26.4& \textbf{\textit{63.1}}& \textbf{\textit{64.2}}& \textbf{\textit{20.0}}& \textbf{\textit{21.2}}& \textbf{\textit{29.0}} & \textbf{\textit{36.6}} & \textbf{\textit{7.7}} & \textbf{\textit{11.4}} & \textbf{\textit{21.2}} & \textbf{\textit{27.5}} & \textbf{\textit{6.8}} & 10.8 & \textbf{\textit{63.7}} & \textbf{\textit{16.1}}\\
% &RelTR$^\bullet$ & 27.1& 51.8& 53.2& 26.9& 28.4 & 24.7 &30.6 &10.8 & 15.4 & 18.6 & 24.1 & 9.2 &13.9 & \textbf{\textit{63.7}} & \textbf{\textit{16.1}}\\
% &RelTR$^\diamondsuit$  & 26.4& 54.3 &55.4& \textbf{\textit{30.3}}& \textbf{\textit{31.8}}& 26.1 & 32.7 & \textbf{\textit{11.9}} & \textbf{\textit{17.2}} & 19.8 & 25.9 & \textbf{\textit{9.7}} & 14.2 & \textbf{\textit{63.7}} & \textbf{\textit{16.1}}\\
\hline \hline
\end{tabular}
\end{adjustbox}
\caption{Comparison with state-of-the-art scene graph generation methods on Visual Genome~\cite{krishna2017visual} test set. These methods are divided into two-stage and one-stage. 
The best numbers in two-stage methods are shown in \textbf{bold}, and the best numbers in one-stage methods are shown in \textbf{\textit{italic}}. 
Models that use  prior knowledge are represented in \textcolor{blue}{blue}, to distinguish them from visual-based models. 
The inference speed (FPS) of different models is tested on the same RTX 2080Ti of batch size 1. 
%{Note that the number of FCSGG is directly taken from \cite{liu2021fully} due to unavailable code.}
}
% We adopt the inference speed published in their paper which is measured on a GTX 1080 Ti.
\label{tab:quantitative_result}
\end{table*}

%%%%%%%%%%%%%%%%%%

Given the triplet cost matrix $\bm{C}_{tri}$, the Hungarian algorithm is executed for the bipartite matching and each ground truth triplet is assigned to a prediction.
However, \texttt{$<$background-no relation-background$>$} should not be assigned to all predictions that do not match the ground truth triplets. 
After several iterations of training, RelTR is likely to output the triplet proposals in four possible ways, as demonstrated in Fig.~\ref{fig:matching}. 
Assigning ground truth to Proposal A and \texttt{$<$background-no relation-background$>$} to Proposal B are two clear cases. 
For Proposal C,  \texttt{$<$background$>$} should not be assigned to the subject due to poor object prediction. 
Furthermore, \texttt{$<$background$>$} should not be assigned to the subject and object of Proposal D due to the fact that  there is a better candidate Prediction A. 
To solve this problem, we integrate an IoU-based assignment strategy in our set prediction loss: For a triplet prediction, if the predicted subject or object label is correct, and the IoU of the predicted box and ground truth box is greater than or equal to the threshold $T$, the loss function does not compute a loss for the subject or object. 
The set prediction loss for triplet detection is formulated as:
%%%%%%%%%%%%%%%%%%
\begin{equation}
\centering
\begin{aligned}
%&\Theta = \vmathbb{0}_{\left\{ \bm{P}^_i= \phi \land IOU>T \land cls \right\}} \\
&L_{sub} = \sum_{i=1}^{N_t} \Theta \left[ L_{cls} + \vmathbb{1}_{\left\{ c_{sub}^i\ne \phi \right\} }L_{box} \right]\\
&L_{obj} = \sum_{i=1}^{N_t} \Theta \left[ L_{cls} + \vmathbb{1}_{\left\{ c_{obj}^i\ne \phi \right\} }L_{box} \right]\\
&L_{triplet}=L_{sub}+L_{obj}+L^{prd}_{cls},
\end{aligned}
\label{eq:loss_triplet}
\end{equation}
%%%%%%%%%%%%%%%%%%
where $L^{prd}_{cls}$ is the cross-entropy loss for predicate classification. $\Theta$ is 0, when \texttt{$<$background$>$} is assigned to the subject/object but the label is predicted correctly and the box overlaps with the ground truth IoU$ \geqslant T$; in other cases, $\Theta$ is 1. The total loss function is computed as:
%%%%%%%%%%%%%%%%%%
\begin{equation}
\centering
\begin{aligned}
&L_{total}=L_{entity}+L_{triplet}.
\end{aligned}
\label{eq:loss_total}
\end{equation}
%%%%%%%%%%%%%%%%%%

\subsection{Post-processing}
\begin{figure}[http]
\centering
\includegraphics[width=1\linewidth]{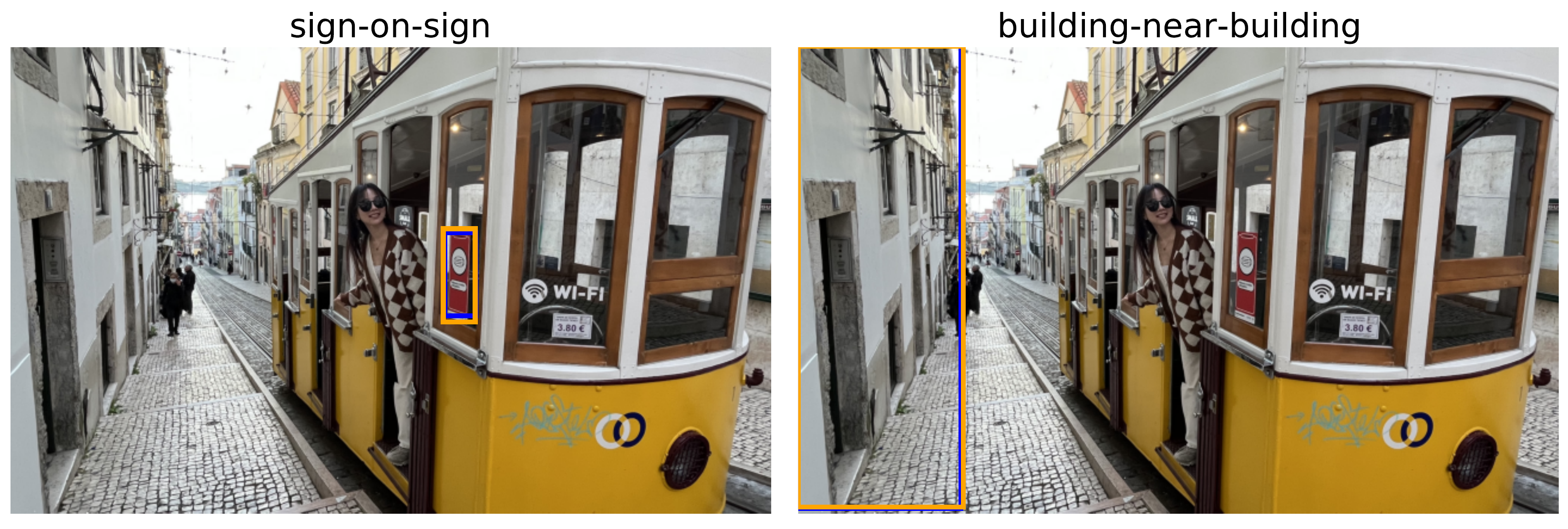}   
\caption{Triplets in which the subject (blue) and object (orange) are the same entity are removed in post-processing. The predicates are usually ambiguous in such cases.} 
\label{fig:duplicate}
\end{figure}

Unlike two-stage methods that organize $N$ entities into $N(N-1)$ subject-object pairs, our method simultaneously detects subjects and objects while predicting a fixed number of triplets.
This results in our approach missing the constraint that the subject and object cannot be the same entity.
It turns out  that our model sometimes outputs a kind of triplet, where the subject and object are the same entity with an ambiguous predicate (see Fig.~\ref{fig:duplicate} for example).
In post-processing, if the subject and object are the same entity (determined by the labels and the bounding boxes' IoU), the triplet is removed. %from the result.
%%%%%%%%%%%%%%%%%%
%%%%%%%%%%%%%%%%%% 

%%%%%%%%%%%%%%%%%% Quantitative Results 
\begin{figure*}[http]
\centering
\includegraphics[width=1\linewidth]{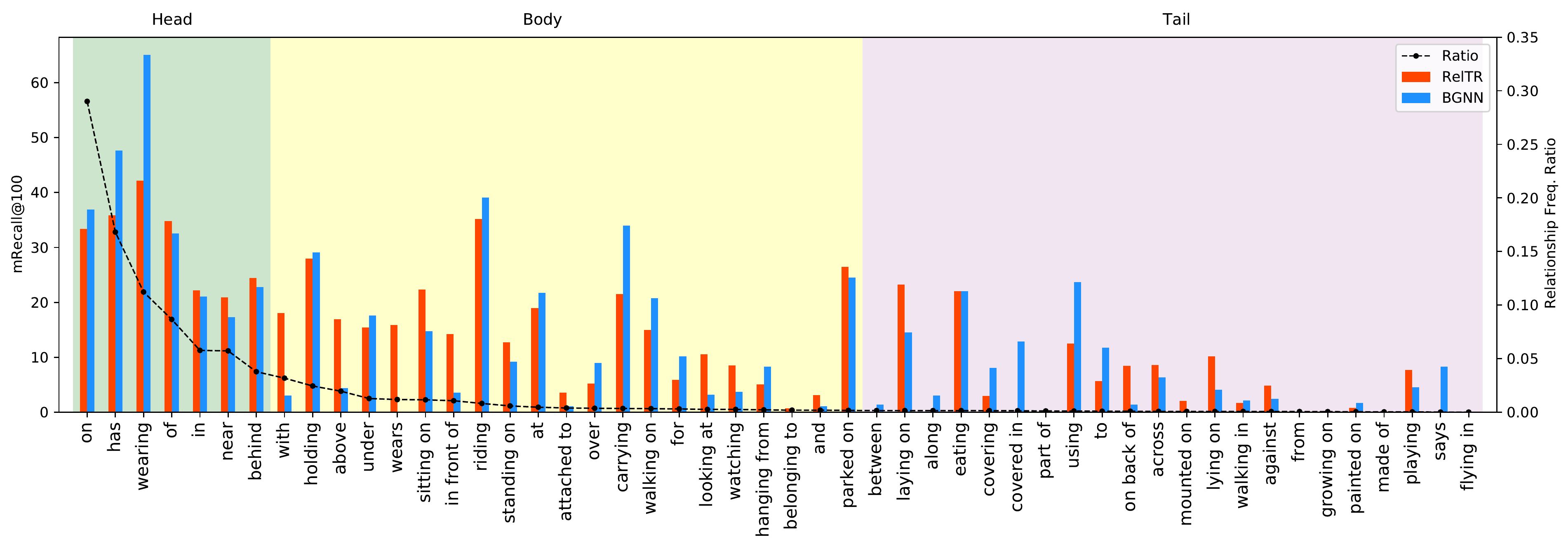}
\caption{SGDET-R$@100$ for each relationship category on VG dataset. Long-tail groups are shown with different colors. RelTR almost always performs better than BGNN~\cite{li2021bipartite} from \texttt{of} to \texttt{in front of}. The standard deviation of R$@100$ are respectively 11.51 (ours) and 14.15 (BGNN). It indicates that RelTR is more unbiased.}
\label{fig:longtail_vg}
\end{figure*}  
%%%%%%%%%%%%%%%%%%

\section{Experiments}
\label{sec:exp}

\subsection{Datasets and Evaluation Settings}
\subsubsection{Visual Genome}
We followed the widely used Visual Genome~\cite{krishna2017visual} split proposed by \cite{xu2017scene}. There are a total of $108k$ images in the dataset with 150 entity categories and 50 predicate categories. $70\%$ of the images are divided into the training dataset and the remaining $30\%$ are used as the test set. 
$5k$ images are further drawn from the training set for validation. There are three standard evaluation settings: 
(1) Predicate classification (\textbf{PredCLS}): predict predicates given ground truth categories and bounding boxes of entities. 
(2) Scene graph classification (\textbf{SGCLS}): predict predicates and entity categories given ground truth boxes. 
(3) Scene graph detection (\textbf{SGDET}): predict categories, bounding boxes of entities and predicates. 
Distinct from two-stage methods, the ground truth bounding boxes and categories of entities cannot be given directly. 
Therefore, we assign the ground truth information to the matched triplet proposals when evaluating RelTR on \textbf{PredCLS}/\textbf{SGCLS}. Recall$@k$ (R$@k$), mean Recall$@k$ (mR$@k$), zero-shot Recall$@k$ (zsR$@k$), no-graph constraint Recall$@K$ (ng-R$@K$), and no-graph constraint zero-shot Recall$@K$ (ng-zsR$@K$) are adopted to evaluate the algorithm performance \cite{lu2016visual,tang2019learning}.
To better estimate the model performance on the imbalanced VG dataset, the relationship categories are split into three groups based on the number of instances in training ~\cite{li2021bipartite}: head ($>10k$), body ($0.5k-10k$) and tail ($<0.5k$).

\subsubsection{Open Images V6}
We conduct experiments on the large-scale Open Images V6 ~\cite{kuznetsova2020open} consisting of $126k$ training images, $5.3k$ test images, and $1.8k$ validation images. 
It involves 288 entity categories and 30 predicate categories. 
We adopt the standard evaluation metrics used in the Open Images Challenge. Recall$@50$, weighted mean average precision (AP) of relationship detection wmAP\textsubscript{$rel$}, and phrase detection wmAP\textsubscript{$phr$} are calculated. 
The final score is computed as: score\textsubscript{$wtd$} = $0.2 {\times}$R$@50$+$0.4 {\times}$wmAP\textsubscript{$rel$} +$0.4 {\times}$wmAP\textsubscript{$phr$}.

\subsubsection{Visual Relationship Detection}
We also validate RelTR on the Visual Relationship Detection (VRD)  dataset~\cite{lu2016visual}, which contains $4k$ training images and $1k$ test images.
R@50 and R@100 in relationship detection and phrase detection are reported, which are used in ~\cite{lu2016visual}.

\subsection{Implementation Details}
%We adopt the same hyperparameters in our experiments on Visual Genome and Open Images. 
For Visual Genome and Open Images, we train RelTR end-to-end from scratch for 150 epochs on 8 RTX 2080 Ti GPUs with AdamW~\cite{loshchilov2017decoupled} setting the batch size to 2 per GPU, weight decay to $10^{-4}$ and clipping the gradient norm$>0.1$.  
The initial learning rates of the Transformer and ResNet-50 backbone are set to $10^{-4}$ and $10^{-5}$ respectively and the learning rates are dropped by 0.1 after 100 epochs.
For small-sized VRD, previous two-stage methods~\cite{zhang2019graphical,lin2020gps} adopt the entity detectors pretrained on ImageNet~\cite{deng2009imagenet} and COCO~\cite{lin2014microsoft}.
Our single-stage method cannot directly utilize these pretrained detectors. 
Instead, we initialize RelTR with Visual Genome pretrained weights, except for the subject, object, and predicate classifiers.
We finetune RelTR on VRD for 100 epochs.
The learning rates for the classifiers are set to $10^{-4}$ and for the other modules are set to $10^{-5}$. 
For all three datasets, we also use auxiliary losses~\cite{al2019character} for the triplet decoder as~\cite{carion2020end,zhu2020deformable} did in the training. 
By default, RelTR has 6 encoder layers and 6 triplet decoder layers. 
The number of triplet decoder layers and the number of entity decoder layers are set to be the same.
The multi-head attention modules with 8 heads in our model are trained with dropout of 0.1. 
For all experiments, the model dimension $d$ is set to 256. 
If not specifically stated, the number of entity queries $N_e$ and coupled queries $N_t$ are respectively set to 100 and 200 while the IoU threshold in the triplet assignment is 0.7. 
For fair comparison, inference speeds (FPS) of all the reported SGG models are evaluated on a single RTX 2080 Ti with the same hardware configuration.
For computing the inference speed (FPS), we average over all the test images, where for each image, the time cost for start timing when an image is given as input and end timing when triplet proposals are output as the inference time.
The time cost for evaluating the whole dataset is not included.

\subsection{Quantitative Results and Comparison}
\subsubsection{Visual Genome}
We compare scores of R$@K$ and mR$@K$, number of parameters, and inference speed on SGDET (FPS) with several two-stage models and two one-stage models \cite{liu2021fully,li2022sgtr} in Table~\ref{tab:quantitative_result}. 
Models that not only use visual appearance but also prior knowledge (e.g. semantic and statistic information) are represented in blue, to distinguish them from visual-based models.
Overall, the two-stage models have higher scores of R$@K$ and mR$@K$ than the one-stage models while they have more parameters and slower inference speed.
This phenomenon also occurs between the models using prior information and visual-based models.
Noted that the performance of the entity detectors in the two-stage models has a significant impact on the model's scores, especially on SGDET.
Our model achieves R$@50=27.5$ and mR$@50=10.8$ on SGDET, which is respectively $5.1$ and $6.2$ points higher than the one-stage model FCSGG~\cite{liu2021fully}. 
RelTR also outperforms SGTR \cite{li2022sgtr} in terms of  R$@50$ on SGDET, while SGTR has higher mR$@50$ due to its graph assembling module. 
Not only that, RelTR has fewer parameters and faster inference speed.
Our model is also competitive compared with recent two-stage models and outperforms state-of-the-art visual-based methods. 
Although the R$@20$/R$@50$ score of RelTR is 2.1/3.5 points lower than that of BGNN~\cite{li2021bipartite}, the performance of RelTR on mR$@50$ is higher. 
Furthermore, RelTR is a light-weight model, which has only 63.7M parameters and an inference speed of 16.6 FPS, ca. 7 times faster than BGNN. 
This allows RelTR to be used in a wide range of practical applications. 
For PredCLS and SGCLS, the ground truth bounding boxes and labels of entities cannot be given to RelTR directly. 
Therefore, we replace the predicted boxes and labels of the matched triplet proposals with ground truth information.
However, it is not possible to capture the exact features of the given boxes by RoIAlign as in two-stage methods.
RelTR uses the features of detected proposals to predict the labels and achieves R$@50=64.2$ and mR$@50=21.2$ on PredCLS while R$@50=36.6$ and mR$@50=11.4$ on SGCLS.

Table~\ref{tab:zeroshot_vg} demonstrates R$@K$, mR$@K$ and zsR$@k$ on SGDET of state-of-the-art methods. 
These methods are divided into unbiased SGG methods and general SGG methods.
Compared with the general models without unbiased learning, RelTR has the best performance on mR$@K$ and zsR$@k$.
zsR$@k$ and mR$@K$ of the two-stage methods with unbiased learning \cite{tang2020unbiased,guo2021general,suhail2021energy,desai2021learning} are improved whereas R$@K$ decreases significantly. 
Our model performs well and is balanced on all three recall metrics. 
Table~\ref{tab:nograph_vg} shows no-graph constraint ng-R$@K$ and ng-zsR$@K$ on SGDET, where multiple predicates are allowed for each subject-object pair. 
%%%%%%%%%%%%%%%%%%
\begin{table}[!htbp]
\centering
\begin{adjustbox}{width=0.49\textwidth}
\begin{tabular}{c|cccccc|c}
 \hline \hline
 \multirow{2}*{Method} & \multicolumn{6}{c|}{SGDET} &  \multirow{2}*{Avg.}\\ &R@20 & R@50& mR@20 & mR@50& zsR@50 & zsR@100 &\\
  \hline
Motifs-TDE\cite{tang2020unbiased}&12.4&16.9&5.8 & 8.2& 2.3& 2.9& 8.1\\
VTransE-TDE\cite{tang2020unbiased}&13.5 & 18.7& 6.3 & 8.6  & 2.0 & 2.7&8.6\\
VCTree-TDE\cite{tang2020unbiased}&14.0 & 19.4& 6.9 & 9.3 & \textbf{2.6} & \textbf{3.2}&9.2\\
%Motifs (BA-SGG)\cite{guo2021general}& 16.8 & 23.0 & 10.7 &13.5 & - & - &-\\
VCTree (BA-SGG)\cite{guo2021general}& 15.8& 21.7 & 10.6 &13.5 & - & - &-\\
%Motifs (EMB)\cite{suhail2021energy}& 24.3 &31.7 & 5.7& 7.7 & 0.1 &0.2 &11.6\\
VCTree-TDE (EMB)\cite{suhail2021energy}& 14.7 & 20.6 & 7.1 &9.7 & 1.6 & 2.7 &9.4\\
DT2-ACBS\cite{desai2021learning}& - & - & \textbf{16.7} &\textbf{22.0} & - & - &-\\
  \hline
Motifs\cite{zellers2018neural}& 21.4&27.2 & 4.2& 5.7  & 0.1& 0.2&9.8\\
VTransE \cite{zhang2017visual}&\textbf{23.0} & \textbf{29.7}& 3.7 & 5.0  & 0.8 & 1.5&10.6\\
VCTree \cite{tang2019learning}& 22.0 & 27.9& 5.2 & 6.9 & 0.2 & 0.7&10.5\\
FCSGG\cite{liu2021fully}&16.1&21.3& 2.7 & 3.6  & 1.0& 1.4&7.7\\
RelTR (ours) &21.2& 27.5& 6.8 & 10.8 & 1.8& 2.4& 11.8\\
% RelTR$^\bullet$ & 18.6 & 24.1 & 9.2 & 13.9 & 2.0 & 2.5 & 11.7\\
% RelTR$^\diamondsuit$ & 19.8 & 25.9 & 9.7 & 14.2 & 1.9& 2.7& \textbf{12.4}\\
\hline \hline
\end{tabular}
\end{adjustbox}
\caption{R$@K$, mR$@K$ and zsR$@k$ performance comparison. The last column is the average of the first six columns. Although the unbiased models have better performance on zsR$@k$, their R$@K$ drops significantly.
Our visual-based model performs balanced and well on the three metrics.}
\label{tab:zeroshot_vg}
\end{table}
%%%%%%%%%%%%%%%%%%
%%%%%%%%%%%%%%%%%%
\begin{table}[!htbp]
\centering
\begin{adjustbox}{width=0.49\textwidth}
\begin{tabular}{c|cccc}
 \hline \hline
 \multirow{2}*{Method} & \multicolumn{4}{c}{SGDET}  \\ 
 & ng-R@50& ng-R@100 & ng-zsR@50& ng-zsR@100 \\
  \hline
Pixels2Graphs \cite{newell2017pixels}&9.7&11.3& - & -\\
KERN \cite{chen2019knowledge}&\textbf{30.9}&35.8&- &- \\
PCPL \cite{yan2020pcpl}&15.2&20.6&- &- \\
GB-NET \cite{zareian2020bridging}&29.3&35.0&- & - \\
Motifs \cite{zellers2018neural}&30.5& 35.8&- & - \\
RelDN \cite{zhang2019graphical}&30.4& \textbf{36.7}& -& - \\
\hline
FCSGG \cite{liu2021fully}& 23.5 &29.2& 1.4 & 2.3 \\
RelTR (ours)  & 30.7& 35.2& \textbf{2.6}& \textbf{3.4}\\
\hline \hline
\end{tabular}
\end{adjustbox}
\caption{Results of no-graph constraint Recall$@k$ (ng-R$@K$) and zero shot Recall$@k$ (ng-zsR$@K$) on Visual Genome. 
}
\label{tab:nograph_vg}
\end{table}
%%%%%%%%%%%%%%%%%%

To further analyze the model performance on imbalanced Visual Genome, we compute mR$@100$ for each relationship group on SGDET in Table~\ref{tab:longtail_group}. 
Our method outperforms the prior works~\cite{lin2020gps,tang2020unbiased,li2021bipartite} on the body group while mR$@100$ on the tail group is similar to the best BGNN ~\cite{li2021bipartite}. 
RelTR achieves the highest mR$@100$ over all relationship categories. 
The results for each relation category are shown in Fig.~\ref{fig:longtail_vg}.
From \texttt{of} to \texttt{in front of}, RelTR almost always performs better than BGNN~\cite{li2021bipartite} while mR$@100$ of the three most frequent predicates are lower. This could explain why R$@k$ of RelTR is not very high but our qualitative results perform well and the relationships in the generated scene graphs are semantically diverse.
%%%%%%%%%%%%%%%%%%
\begin{table}[!htbp]
\centering
\begin{adjustbox}{width=0.49\textwidth}
\begin{tabular}{c|c|ccc}
 \hline \hline
Method  & SGDET-mR@100 & Head& Body& Tail\\
  \hline 
GPS-NET\cite{lin2020gps}  & 9.8 & 30.8& 8.5& 3.9\\
VCTree-TDE\cite{tang2020unbiased}  & 11.1 & 24.7& 12.2& 1.8\\
BGNN \cite{li2021bipartite} & 12.6 & \textbf{34.0}& 12.9& \textbf{6.0}\\
%SGTR \cite{li2022sgtr}  & \textbf{20.1}& 21.7 &\textbf{21.6} & \textbf{17.1}\\
RelTR  & 12.6& 30.6& 14.4& 5.0\\
% RelTR $^\bullet$ & 15.4& -& -& -\\
% RelTR $^\diamondsuit$ & 16.6& 28.3& 19.4& 10.2\\
\hline \hline
\end{tabular}
\end{adjustbox}
\caption{SGDET-mR$@100$ for the head, body and tail groups which are partitioned according to the number of relationship instances in the training set.} 
\label{tab:longtail_group}
\end{table}
%%%%%%%%%%%%%%%%%%
%%%%%%%%%%%%%%%%%% Quantitative Results 
\begin{figure*}[!htbp]
\centering
\includegraphics[width=1\linewidth]{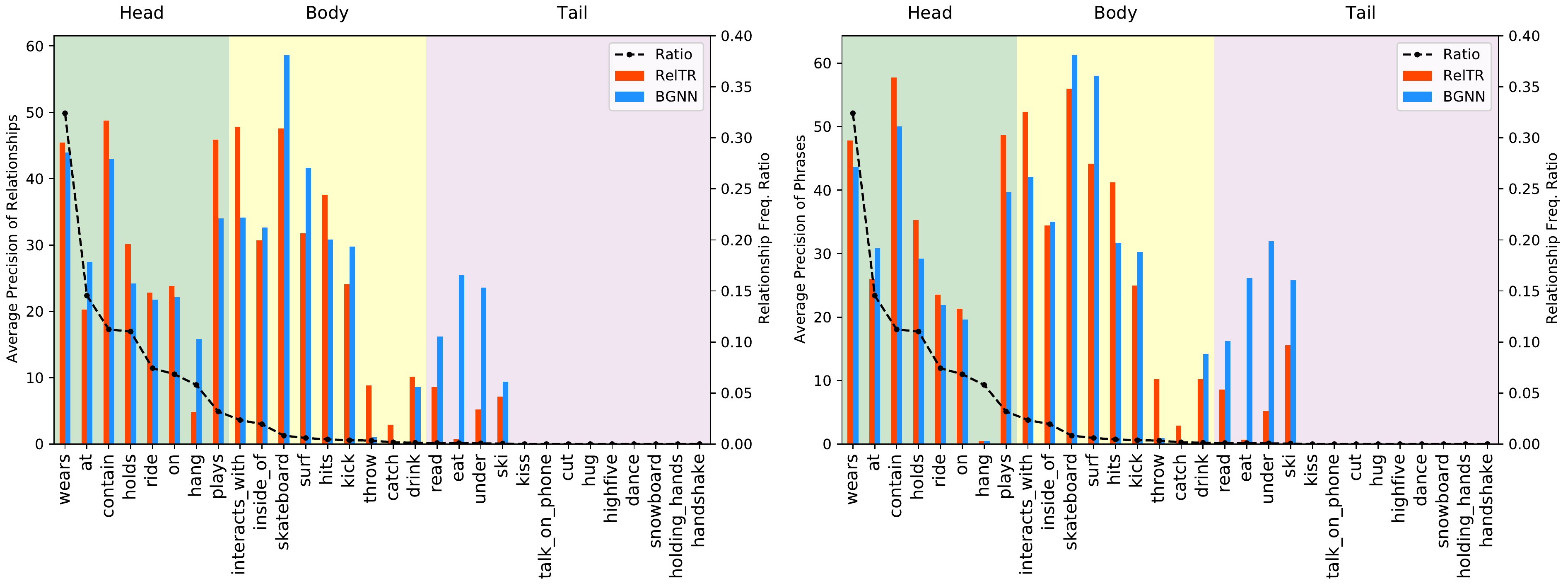}
\caption{Average precision of relationships and phrases for RelTR and BGNN on Open Images V6. The distribution of relationships in the test set is shown with the black dash line. The average precision of relationships of RelTR is higher than BGNN for 7 of the top-10 high frequency predicates while BGNN generally performs better than RelTR for the low frequency predicates (\texttt{skateboard} to \texttt{ski}). We conjecture that it is attributed to prior knowledge used in BGNN. The overall trend of AP$_{phr}$ is the same as AP$_{rel}$ except \texttt{hang}.}
\label{fig:map_rel_phr}
\end{figure*}  
%%%%%%%%%%%%%%%%%%
\subsubsection{Open Images V6} 
We train RelTR on the Open Images V6 dataset and compare it with other two-stage methods and another one-stage method SGTR \cite{li2022sgtr}, as shown in Table~\ref{tab:quantitative_result_oi}. 
Although R$@50$ of RelTR is 3.68 points lower than the best two-stage method VCTree~\cite{tang2019learning}, RelTR has the higher wmAP\textsubscript{$rel$} (0.58 points higher than BGNN~\cite{li2021bipartite}) and wmAP\textsubscript{$phr$} (3.15 points higher than VCTree).
The final weighted score of RelTR is 1.02 points higher than the best two-stage model VCTree.
The one-stage method SGTR performs slightly better on wmAP\textsubscript{$rel$} and wmAP\textsubscript{$phr$}, whereas its R$@50$ is low compared to the other methods.
The inference speed of RelTR is 16.3 FPS, ca. 6 and 4 times faster than BGNN and SGTR, respectively. 
%%%%%%%%%%%%%%%%%%
\begin{table}[!htbp]
\centering
\begin{adjustbox}{width=0.47\textwidth}
\begin{tabular}{c|cccc|c}
 \hline \hline
Method  & R@50 $\uparrow$ & wmAP$_{rel}$ $\uparrow$& wmAP$_{phr}$ $\uparrow$& score$_{wtd}$ $\uparrow$& FPS $\uparrow$\\
  \hline 
RelDN \cite{zhang2019graphical} & 73.08 &32.16 & 33.39 & 40.84 & 5.3 \\
VCTree \cite{tang2019learning} & \textbf{75.34}&33.21 & 34.31 & 41.97 & 1.9\\
G-RCNN \cite{yang2018graph} & 74.51&33.15 & 34.21 & 41.84  & -\\
Motifs \cite{zellers2018neural} & 71.63&29.91 & 31.59 & 38.93 & 7.4\\
GPS-NET \cite{lin2020gps} & 74.81&32.85 & 33.98 & 41.69 & -\\
BGNN \cite{li2021bipartite} & 74.98& 33.51 & 34.15 & 41.69 & 2.9 \\ \hline 
SGTR \cite{li2022sgtr} & 59.91 & \textbf{36.98} & \textbf{38.73} & 42.28 & 3.8 \\ 
RelTR (ours) & 71.66& 34.19 & 37.46 & \textbf{42.99} & \textbf{16.3}\\
\hline \hline
\end{tabular}
\end{adjustbox}
\caption{Comparison with other state-of-the-art methods on the Open Images V6~\cite{kuznetsova2020open} test set. The numbers of these methods are taken from \cite{li2021bipartite, li2022sgtr}.} 
\label{tab:quantitative_result_oi}
\end{table}
%%%%%%%%%%%%%%%%%%

To further demonstrate the performance of RelTR, we compare the average precision (AP) of relationships and phrases for RelTR and BGNN~\cite{li2021bipartite} (see Fig.~\ref{fig:map_rel_phr}) with Open Images V6.
% The license of Open Images V6 is CC BY 4.0. 
% Note that the annotations are licensed by Google LLC under CC BY 4.0 license and the images have a CC BY 2.0 license.
Although R$@50$ of RelTR is lower, RelTR outperforms BGNN on the weighted mean AP of relationships and phrases. 
The distribution of relationships in the Open Images V6 test set is also shown with the black dash lines. 
There are 9 predicates (\texttt{kiss} to \texttt{handshake}) that do not appear in the test set. 
The average precision of relationships AP$_{rel}$ and AP$_{phr}$ of RelTR are higher than BGNN for 7 of the top-10 high frequency predicates.
For the low frequency predicates (\texttt{skateboard} to \texttt{ski}), BGNN generally performs better than RelTR.
We conjecture that it is attributed to prior knowledge used in BGNN.

\subsubsection{Visual Relationship Detection}
Table~\ref{tab:vrd} shows the comparison of RelTR with other state-of-the-art methods on the VRD dataset.
All the models are two-stage methods using pretrained entity detectors except our RelTR.
In order to obtain promising results for RelTR with little training data, we initialize RelTR with Visual Genome pre-trained weights and fine-tune the subject, object, and predicate classifiers.
RelTR outperforms the other two-stage scene graph generation methods in both relationship detection and phrase detection.
%%%%%%%%%%%%%%%%%%
\begin{table}[!htbp]
\centering
\begin{adjustbox}{width=0.49\textwidth}
\begin{tabular}{c|cc|cc}
 \hline \hline
 \multirow{2}*{Method} & \multicolumn{2}{c|}{Relationship Detection} &  \multicolumn{2}{c}{Phrase Detection} \\ 
 & R@50& R@100 & R@50& R@100 \\
  \hline
VTransE\cite{zhang2017visual}&19.4&22.4&14.1 & 15.2\\
ViP-CNN\cite{li2017vip}&17.3&20.0&22.8 & 27.9 \\
VRL\cite{liang2017deep}&18.2 & 20.8& 21.4 & 22.6 \\
KL distilation\cite{yu2017visual}&19.2 & 21.3& 23.1 & 24.0 \\
MF-URLN\cite{zhan2019exploring}&23.9 & 26.8& 31.5 & 36.1 \\
Zoom-Net\cite{yin2018zoom}&18.9 & 21.4& 24.8 & 28.1 \\
RelDN\cite{zhang2019graphical}&25.3 & 28.6& 31.3 & 36.4 \\
GPS-Net\cite{lin2020gps}&27.8 & 31.7& 33.8 & 39.2 \\
\hline
RelTR (ours)& \textbf{29.2}&\textbf{32.2}&\textbf{34.5}&\textbf{39.8}\\
\hline \hline
\end{tabular}
\end{adjustbox}
\caption{Comparison with other two-stage approaches on the VRD dataset in relationship and phrase detection. 
}
\label{tab:vrd}
\end{table}

%%%%%%%%%%%%%%%%%%
% \revise{
\subsubsection{Long-tailed Techniques}
To demonstrate the compatibility of our visual-based model with long-tailed techniques, we implement two different techniques for RelTR, namely bi-level resampling \cite{li2021bipartite, gupta2019lvis,li2022sgtr} and logit adjustment \cite{menon2021longtail, teng2022structured}.
We validate two approaches on the Visual Genome dataset, where the distribution of predicate classes is imbalanced.
R$@20/50$, mR$@20/50$, and mR$@100$ for the head, body, and tail groups of SGDET are demonstrated in Table~\ref{tab:long_tail}.
When implementing the bi-level resampling strategy, our model achieves higher mR$@20$ and mR$@50$ scores; however, there is a decrease in R$@20$ and R$@50$ performance. In contrast, RelTR with the logit adjustment demonstrates better performance. The mR$@50$ score improves to 14.2, with a minor drop of 1.6 in R$@50$.
Both techniques can improve the inference performance for the relationship classes of the body and tail groups.
The results show that our model has the potential to be extended to an unbiased scene graph generation approach.
 % }

 \begin{table}[!htbp]
\centering
\begin{adjustbox}{width=0.49\textwidth}
\begin{tabular}{c|cccc|ccc}
 \hline \hline
Method  & R@20 &R@50 & mR@20 &mR@50 & Head& Body& Tail\\
  \hline 
RelTR  & 21.2& 27.5 & 6.8 & 10.8 & 30.6& 14.4& 5.0\\
RelTR$+$RS & 18.6 & 24.1 & 9.2 & 13.9 & 29.1 & 17.3 & 10.5\\
RelTR$+$LA & 19.8 & 25.9 & 9.7 & 14.2 & 28.3& 19.4& 10.2\\
\hline \hline
\end{tabular}
\end{adjustbox}
\caption{We implement two long-tail techniques for RelTR, respectively the bi-level resampling (RS) \cite{li2021bipartite, gupta2019lvis, li2022sgtr} and the logit adjustment (LA) \cite{menon2021longtail, teng2022structured}.
The results show that RelTR is  compatible with these long-tailed techniques and the model performance in predicting low-frequency predicates is significantly improved.} 
\label{tab:long_tail}
\end{table}

\subsection{Ablation Studies}
In the ablation studies, we consider how the following aspects influence the final performance.
All the ablation studies are performed with Visual Genome dataset~\cite{krishna2017visual}.

\subsubsection{Number of Layers} 
The feature encoder layer and triplet decoder layer have different effects on the performance, size and inference speed.
%There are two different layers, the encoder layer and decoder layer, which have different effects on the performance, size and inference speed. 
When the number of encoder layers varies, we keep the number of triplet decoder layers always 6, and vice versa. 
When there is no encoder layer, the decoder reasons about the feature map without context and R$@50$ drops by 4.2 points significantly (see Table~\ref{tab:alation_layer}). 
Adding an encoder layer brings fewer parameters compared to adding a triplet decoder layer. 
Because the decoder is indispensable for scene graph generation, the minimum number of triplet decoder layers in our experiment is set to 3. 
%At this point the inference speed is incredible, predicting more than 20 images per second. 
When the number of triplet decoder layers is increased to 6, the improvement of R$@20$, R$@50$ and R$@100$ are obvious. 
In contrast, there is a small decrease in performance when the number of triplet decoder layers is increased to 9. We conjecture that this may be caused by overfitting.
%%%%%%%%%%%%%%%%%%
\begin{table}[!htbp]
\centering
\begin{adjustbox}{width=0.49\textwidth}
\begin{tabular}{cc|ccc|c|c}
 \hline \hline
 \multicolumn{2}{c|}{Layer Number} & \multicolumn{3}{c|}{SGDET} & \multirow{2}*{\#params(M)} & \multirow{2}*{FPS} \\ Encoder& Triplet Decoder &R@20 &R@50 &R@100 & & \\
  \hline 
0& 6&17.6 & 23.3 & 27.1 & \textbf{55.8} & \textbf{18.0}\\
3& 6&20.5 & 26.6 & 29.5 & 59.7 & 17.1\\
9& 6&\textbf{21.4} & \textbf{27.7} & \textbf{30.8} & 67.6 & 15.5\\
\hline
6& 6&21.2 & 27.5 & 30.7 & 63.7 & 16.1\\
\hline
6& 3&19.5 & 25.9 & 29.8 & 48.7 & 19.6\\
6& 9&21.0 & 27.1 & 30.1 & 78.7 & 13.8\\
\hline \hline
\end{tabular}
\end{adjustbox}
\caption{Impact of the number of encoder and decoder layers on the performance, model size and inference speed.} 
\label{tab:alation_layer}
\end{table}
%%%%%%%%%%%%%%%%%%

\subsubsection{Module Effectiveness}
\label{sec:module_effectiveness}
To verify the contribution of each module to the overall effect, we deactivate different modules and the results are shown in Table~\ref{tab:ablation_module}. 
We first ablate the entire triplet decoder (first row) and combine the top $64$ confident entity proposals provided by the entity decoder into $64\times63$ triplet proposals as a two-stage method. 
The feature vectors are concatenated and a 3-layer perceptron is used to predict the relationships. 
This can also be seen as a simple visual-based baseline with DETR~\cite{carion2020end} as the detector. 
Without the triplet decoder, R$@50$ score drops to $18.3$ due to the simplicity of the model.
It indicates that only visual information is used to predict relationships, which is a challenge even for two-stage methods.

\begin{figure}[!t]
\centering
\includegraphics[width=0.99\linewidth]{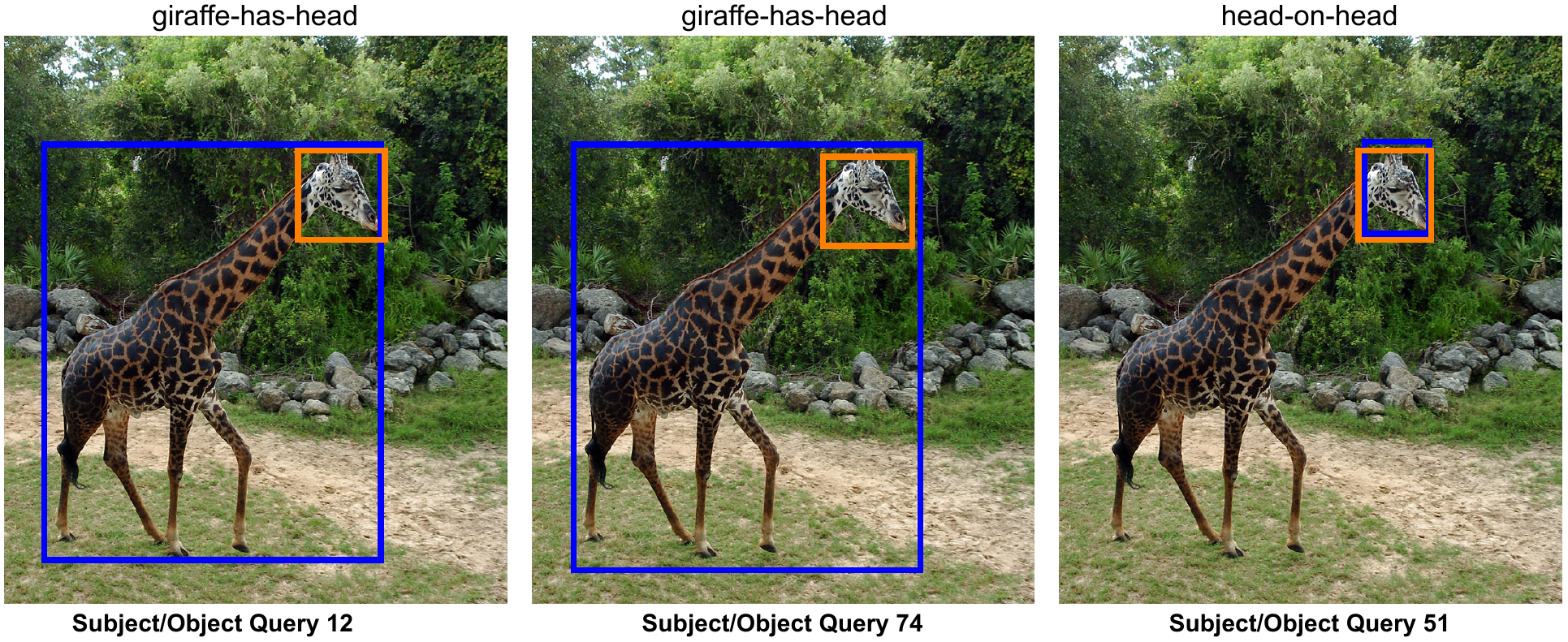}
\caption{Triplet proposals when only DVA modules are activated. Since the subject and object queries are unaware of each other, the $12$-th and $74$-th triplet proposals are duplicated, while the 51-th proposal is semantically meaningless. CSA can suppress these failures.}
\label{fig:ablation}
\end{figure}

To demonstrate the characteristics of each attention module in RelTR, we first activate only the coupled self-attention (CSA),  decoupled visual attention (DVA), and decoupled entity attention (DEA), respectively.
When only CSA is activated  (second row), the model is unable to detect relationships because in the absence of cross-attention, RelTR does not actually receive any visual appearance.
The model can generate normal quality scene graphs when DVA or DEA is integrated.
Using only DVA (third row) is more effective than using only DEA (fourth row) since DVA modules infer visual relationships directly from fine-grained image features.
However, without the support of CSA, the subject and object queries of all triplet proposals are independent and mutually unaware, which leads to multiple triplet proposals linking to the same relationship, or triplets in which the subject and object are the same entity (see Fig.~\ref{fig:ablation}).

Although the triplet decoder is not yet complete, the main modules CSA and DVA (fifth row) have shown excellent performance.
The model parameters are $43\%$ more than the simple baseline, but the model can predict up to $77\%$ of the baseline inference speed (FPS) due to the sparse graph generation method.
In contrast, activating both CSA and DEA has worse performance, but faster inference speed, since only the coarse-grained entity representations are used to generate a scene graph.
Table~\ref{tab:ablation_module} also demonstrates that DEA helps the model to predict higher quality subjects and objects, and increase R$@50$ by 0.6.
In comparison, the improvement offered by the mask head is limited.
We hypothesize that the spatial features are already implicit encoded in the visual features generated by DVA modules.
% We activate the Coupled Self-Attention (CSA) and Decoupled Visual Attention (DVA) simultaneously since they are indispensable to each other (second row).
% Although the triplet decoder is not yet complete, the main modules CSA and DVA have shown their excellent performance. 
% The model parameters are $43\%$ more than the simple baseline, but the model can predict up to $77\%$ of the baseline inference speed (FPS) due to the sparse graph generation method.
% Then we ablate Decoupled Entity Attention (DEA) and the mask head for the attention heat maps from the framework. 
% Table~\ref{tab:ablation_module} demonstrates that DEA modules help the model to predict higher quality subjects and objects, and increase R$@50$ by 0.7 (with the mask head).
% In comparison, the improvement offered by the mask head is very limited.
% We hypothesize that the spatial features are already implicit encoded in the visual features generated by DVA modules.
%%%%%%%%%%%%%%%%%%
\begin{table}[!htbp]
\centering
\begin{adjustbox}{width=0.49\textwidth}
\begin{tabular}{cccc|cccc|c|c}
 \hline \hline
 \multicolumn{4}{c|}{Ablation Setting} & \multicolumn{4}{c|}{SGDET} & \multirow{2}*{\#params(M)} & \multirow{2}*{FPS} \\
 CSA & DVA & DEA & Mask &R@20 &R@50 &mR@20 &mR50 & & \\
  \hline 
\xmark& \xmark& \xmark& \xmark&12.0 & 18.3 &3.5 &5.9  & \textbf{41.5} & 22.0\\
\hline
\cmark& \xmark& \xmark&\xmark & 1.1 & 3.9& 0.3&0.5& 43.6&\textbf{22.1}\\
\xmark& \cmark& \xmark&\xmark & 16.3&20.9&5.0&7.5&57.8&19.6\\
\xmark& \xmark& \cmark&\xmark & 15.0&19.1&4.8&6.9&57.8&20.3\\
\hline
\cmark& \cmark& \xmark&\xmark & 20.6&26.6&6.4&9.6&59.3&17.7\\
\cmark& \xmark& \cmark&\xmark & 17.7&22.2&5.9&8.7&59.3&19.4\\
\xmark& \cmark& \cmark&\xmark & 16.5&21.1&5.1&7.4&60.9&17.5\\
\cmark& \cmark& \cmark&\xmark & 21.0 &27.2&6.4&10.0&62.5&16.7\\
\hline
\cmark& \cmark& \cmark&\cmark & \textbf{21.2}&\textbf{27.5}&\textbf{6.8}&\textbf{10.8} &63.7&16.1\\
\hline \hline
\end{tabular}
\end{adjustbox}
\caption{Coupled self-attention (CSA), decoupled visual attention (DVA), decoupled entity attention (DEA), and the mask head (Mask) for the attention heat maps are isolated separately from the framework. 
The first row indicates that the entire triplet decoder is deactivated and the model can be seen as a simple visual-based baseline with DETR as the detector. 
\xmark $ $ denotes the module is ablated.}
\label{tab:ablation_module}
\end{table}
%%%%%%%%%%%%%%%%%%

\subsubsection{Threshold in Set Prediction Loss}
\begin{figure}[http]
\centering
\includegraphics[width=0.75\linewidth]{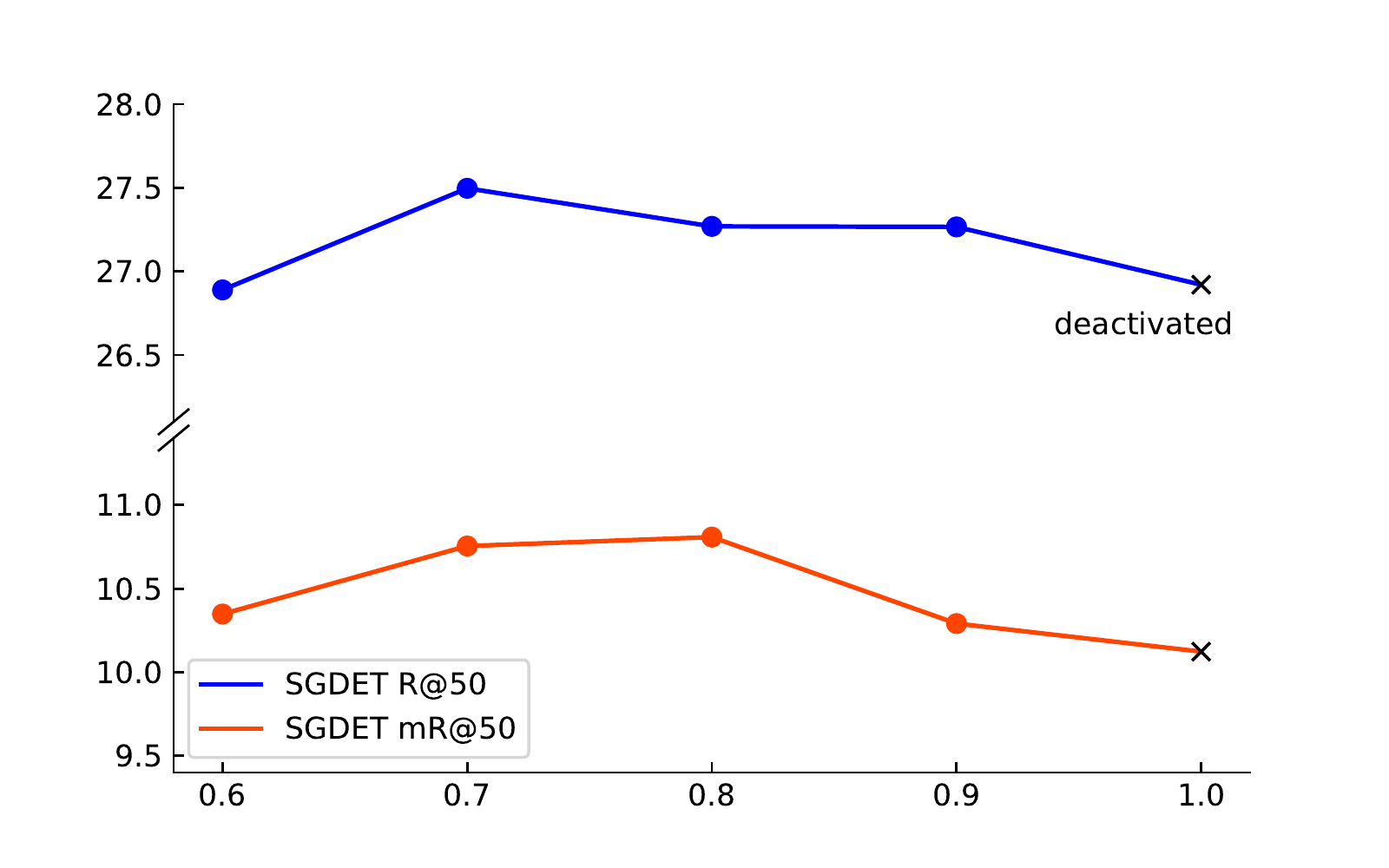}
\caption{$T$-R$@50$ and $T$-mR$@50$ curve on SGDET. $\times$ indicates that the IoU-based assignment strategy is deactivated.} %The values are evaluated on the test set while trained models are sampled based on the validation set.} 
\label{fig:drop_thresh}
\end{figure}

The IoU threshold $T$ of the IoU-based assignment strategy in the set prediction loss for triplet detection is varied from $0.6$ to $1$. 
Since a prediction box overlaps with the ground truth box of IoU$=1$ is almost impossible in practice, the strategy can be considered as deactivated when $T=1$.
Two curves, namely $T$-R$@50$ and $T$-mR$@50$ on SGDET, are shown in Fig.~\ref{fig:drop_thresh}. 
When our assignment strategy is deactivated ($T=1$), the model performs the worst. 
As $T$ increases from 0.7 to 1, the overall trend of the two curves is decreasing. 
This is more evident for the  $T$-mR$@50$ curve.
%%%%%%%%%%%%%%%%%%
%%%%%%%%%%%%%%%%%%
%%%%%%%%%%%%%%%%%%
\begin{figure}[http]
\centering
\includegraphics[width=0.75\linewidth]{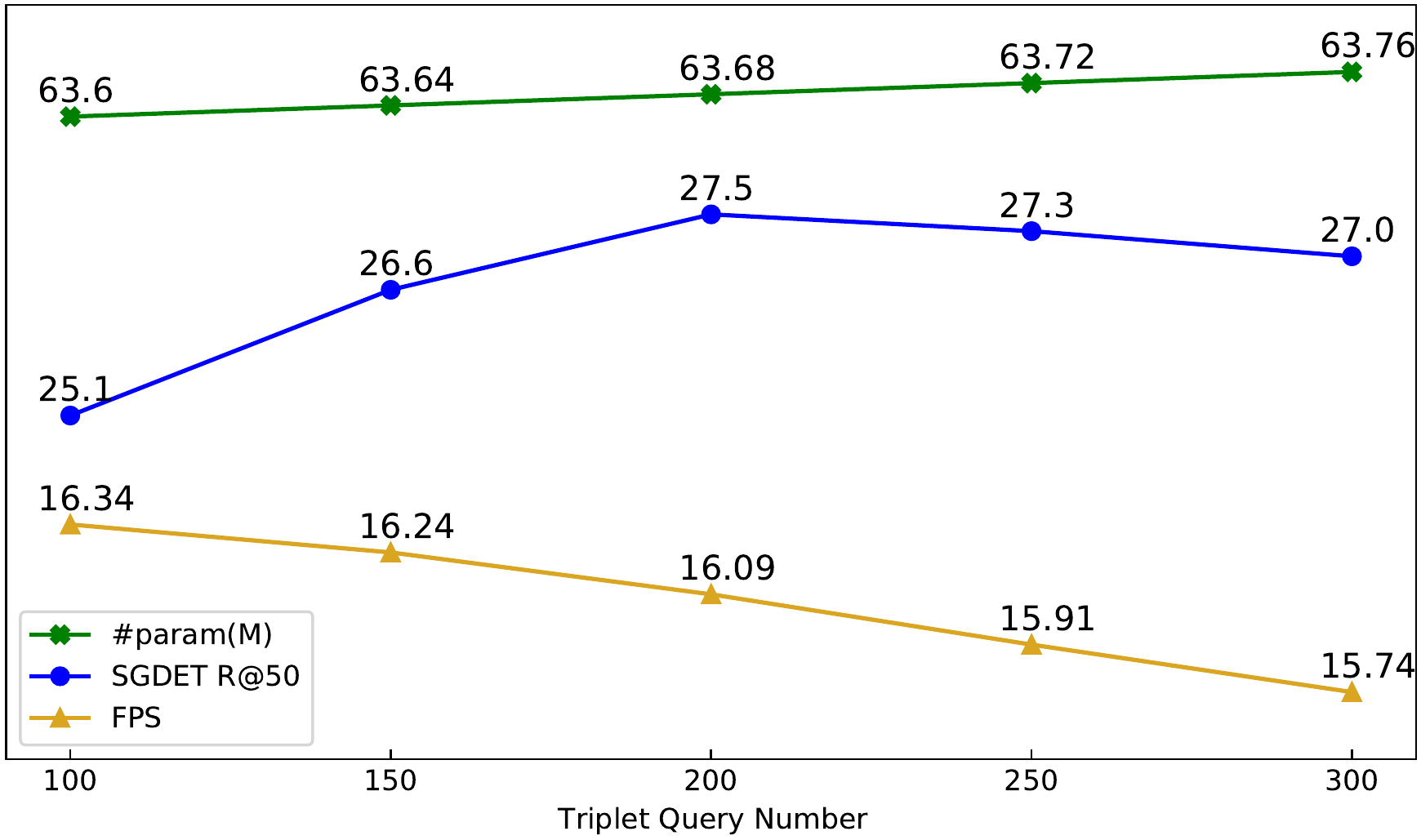}
\caption{Changes in the parameter number, performance and FPS as the triplet number $N_t$ varies.} 
\label{fig:query_num}
\end{figure} 
%%%%%%%%%%%%%%%%%% 

%%%%%%%%%%%%%%%%%%
\begin{figure*}[http]
\centering  
 \includegraphics[width=0.99\linewidth]{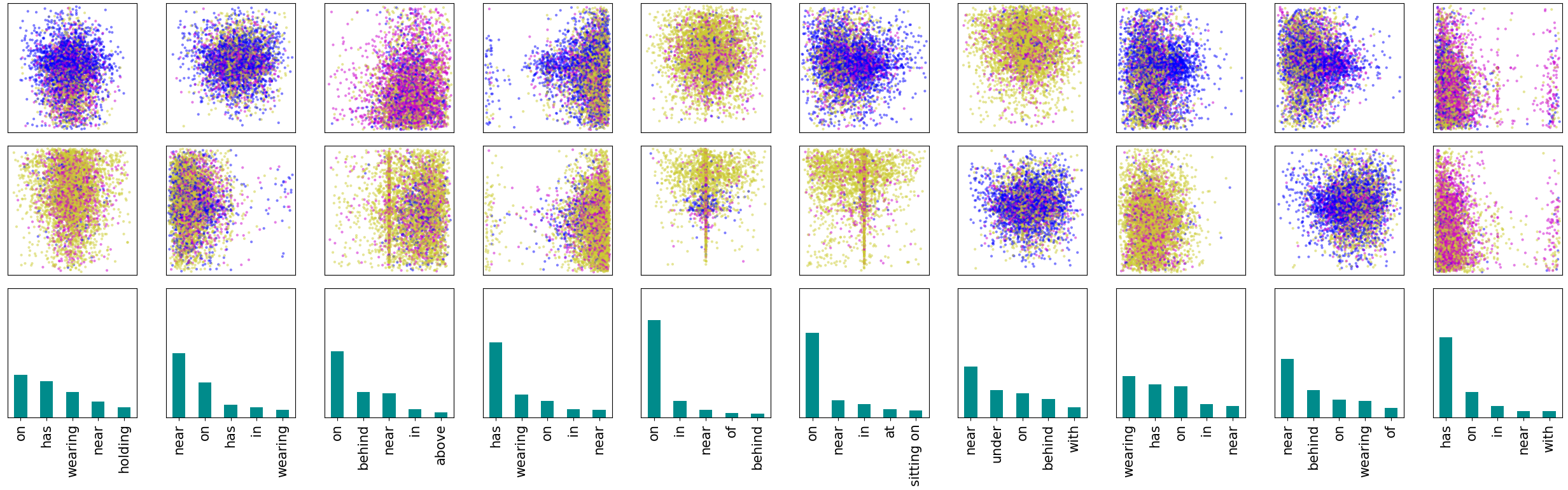}
\caption{Predictions on 5000 images from Visual Genome test set are presented for 10 coupled subject and object queries.
The size of all images is normalized to $1\times1$, with each point in the first and second rows representing the box center of a subject and an object in a prediction respectively. 
Different point colors denote different entity super-categories: (1) blue for humans (child, person and woman etc.) (2) plum for things that exist in nature (beach, dog and head etc.) (3) yellow for man-made objects  (cup, jacket and table etc.). The corresponding distributions of top-5 predicate are shown in the third row.}
\label{fig:triplet_vector}
\end{figure*}  
%%%%%%%%%%%%%%%%%%

\subsection{Analysis on Subject and Object Queries}
Distinct from the two-stage methods which output $N$ object proposals after NMS and then label $N(N-1)$ predicates, RelTR predicts $N_t$ triples directly by $N_t$ subject and object queries interacting with an image. 
We trained the model on Visual Genome using different $N_t$. Fig.~\ref{fig:query_num} shows that as the number of coupled subject and object queries increases linearly, the parameters of the model increase linearly whereas the inference speed decreases linearly. 
However, the performance varies non-linearly and the best performance is achieved when $N_t=200$ for the Visual Genome dataset.
Too many queries generate many incorrect triplet proposals that take the place of correct proposals in the recall list.

\begin{figure}[http]
\centering
\includegraphics[width=1\linewidth]{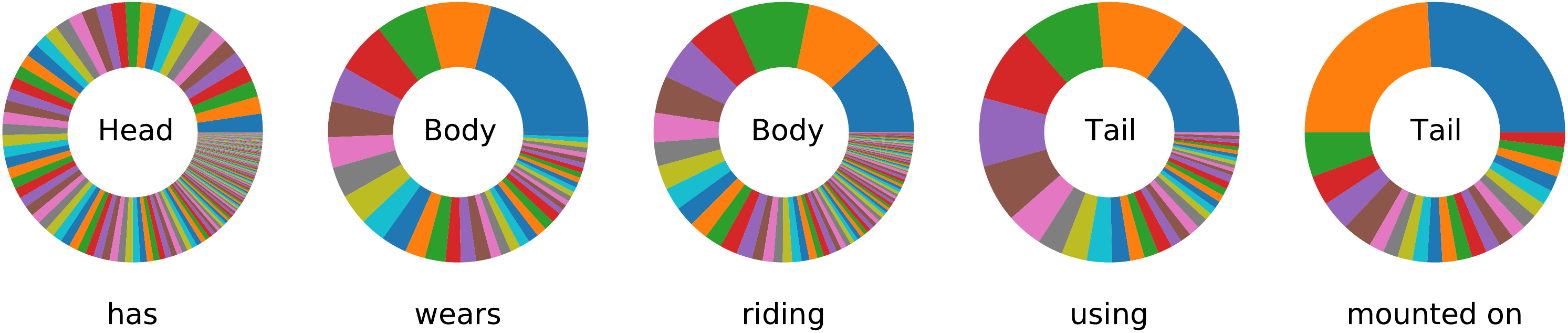}
\caption{Query distribution of the triplets with \texttt{has} (from Head), \texttt{wears}, \texttt{riding} (from Body) \texttt{using} and \texttt{mounted on} (from Tail) in the predictions on 5000 images from Visual Genome test set. Note that the same color in different pie charts does not mean the same query.} 
\label{fig:5pie}
\end{figure}

To explore how RelTR infers triplets with the coupled subject and object queries, we collect predictions from a random sample of 5000 images from Visual Genome test set. 
We visualize the predictions for 10 out of total 200 coupled queries. Fig.~\ref{fig:triplet_vector} shows the spatial and class distributions of subjects and objects, as well as the class distribution of top-5 predicates in the 5000 predictions of 10 coupled subject and object queries. 
It demonstrates that different coupled queries learn different patterns from the training data, and attend to different classes of triplets in different regions at the inference. 
We also select five predicates: \texttt{has} (from Head), \texttt{wears}, \texttt{riding} (from Body) \texttt{using} and \texttt{mounted on} (from Tail) and count which queries are more inclined to predict these predicates. 
As shown in Fig.~\ref{fig:5pie}, the query distribution of \texttt{has} is smooth.
This indicates that all queries are able to predict high frequency relationships.
For predicates in Body and Tail groups, there are some queries that are particularly good at detecting them.
For example, 21\% of the triplets with the predicate \texttt{wears} are predicted by Query 115, while half of the triplets with the predicate \texttt{mounted on} are predicted by Query 107 and 105.

%%%%%%%%%%%%%%%%%%
%%%%%%%%%%%%%%%%%% 

%%%%%%%%%%%%%%%%%%
\begin{figure*}[http]
\centering
\includegraphics[width=0.89\linewidth]{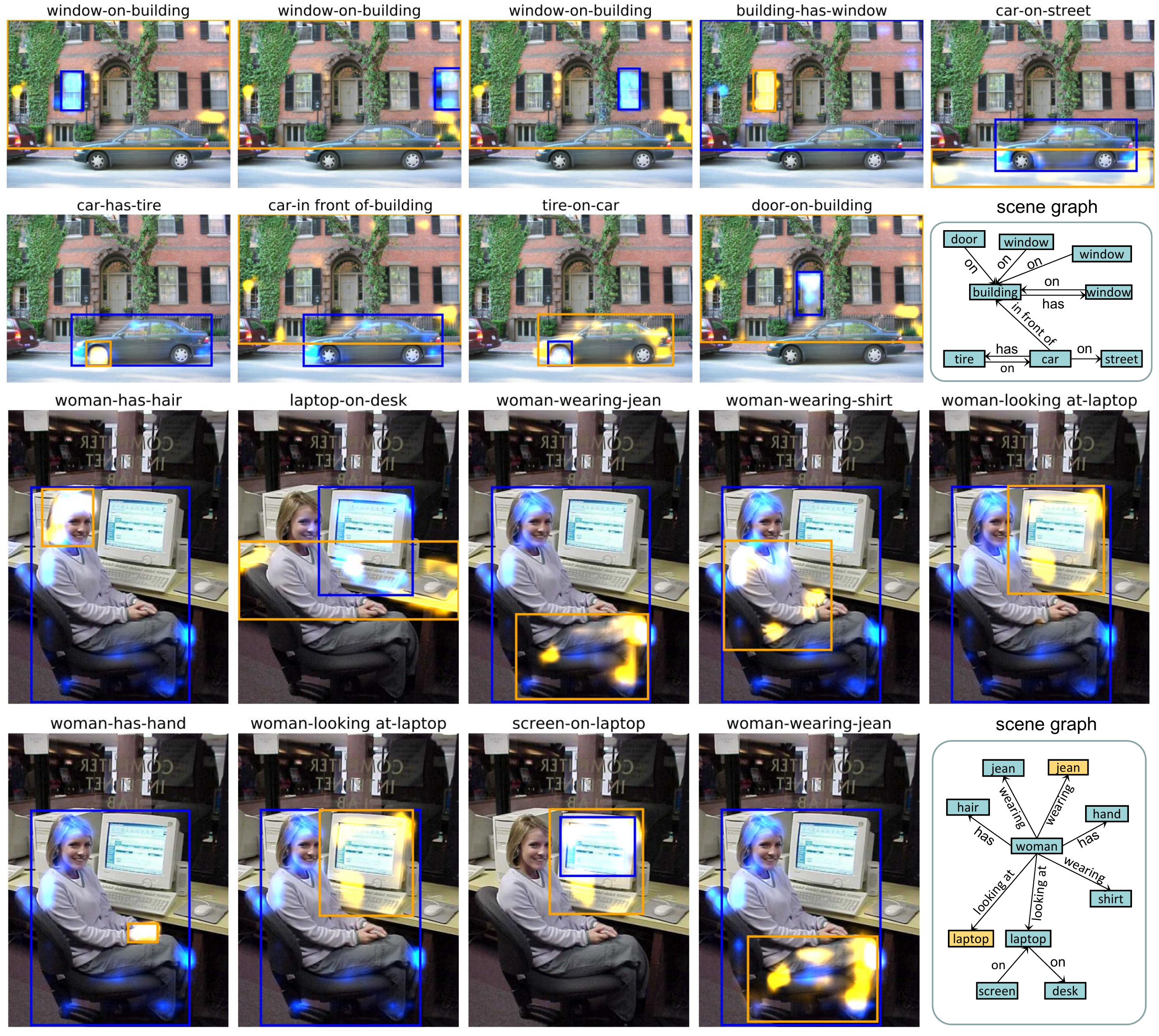}
\caption{Qualitative results for scene graph generation of Visual Genome dataset. The top-9 relationships with confidence and the generated scene graph are shown. Boxes and attention scores of subjects are colored with blue while objects with orange. The orange vertices in the generated scene graph indicate the predictions are duplicated. The \texttt{computer} is classified as \texttt{laptop} in the second image since there is no \texttt{computer} class but only \texttt{laptop} class in the used VG-150 split~\cite{xu2017scene}. Compared with the ground truth annotations in Fig.~\ref{fig:vg_gt}, the predictions of RelTR are diverse. Although sometimes RelTR cannot label very difficult relationships correctly (e.g. \texttt{looking at}), the results demonstrate that the generated scene graphs are of high quality. }
\label{fig:vg_qualitative}
\end{figure*}  
%%%%%%%%%%%%%%%%%%

\subsection{Qualitative Results}
\begin{figure*}[http]
\centering
\includegraphics[width=0.73\linewidth]{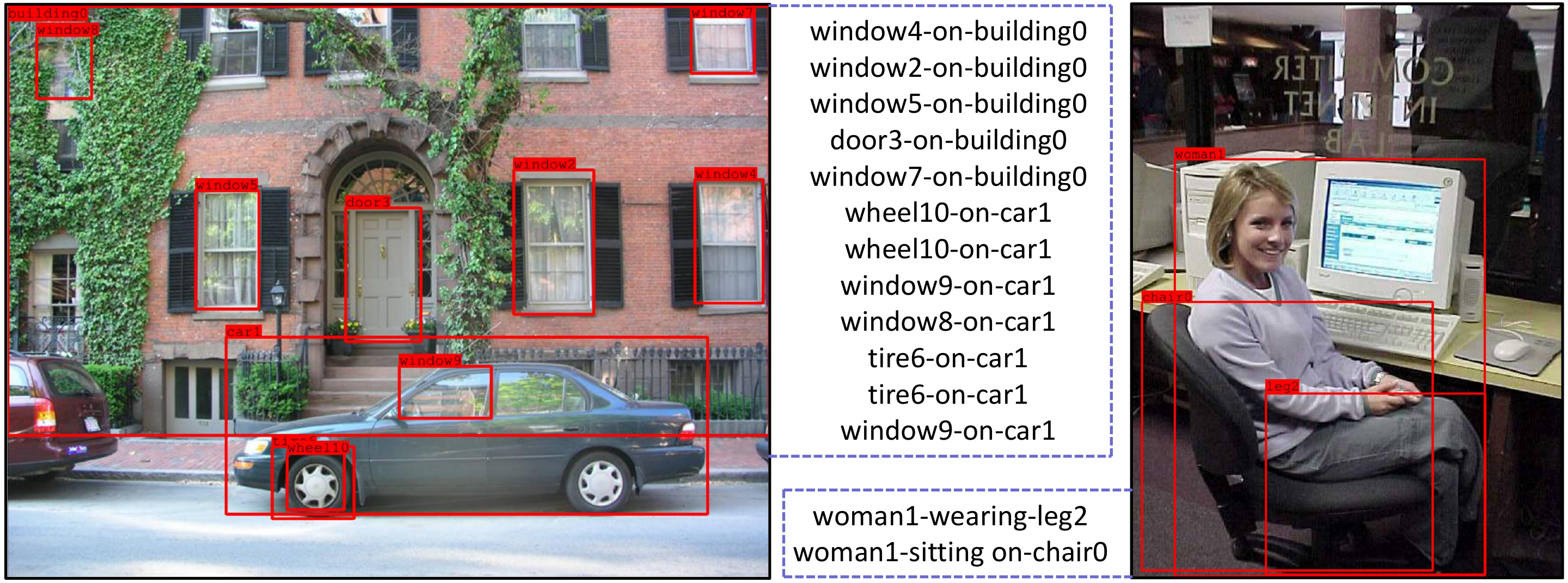}
\caption{Ground truth annotations of the two images in Fig.~\ref{fig:vg_qualitative} from Visual Genome dataset. For brevity, only the bounding boxes of the entities that appear in the annotated triplets are shown with red. 
All entities are numbered to distinguish between entities of the same class. 
There are two errors in the ground truth annotations: \texttt{$<$window8-on-car1$>$} in the first image and \texttt{$<$woman1-wearing-leg2$>$} in the second image. 
There could be duplicate triplets in the ground truth (e.g. \texttt{$<$wheel10-on-car1$>$} in the first image). For the first image, only the relationships with the predicate \texttt{on} are labeled while for the second image, the relationships such as \texttt{$<$woman1-wearing-shirt$>$} are omitted.
These biases in the ground truth annotations lead to the low score of R$@K$, the other SGG models also suffer from this problem. }
\label{fig:vg_gt}
\end{figure*}

Fig.~\ref{fig:vg_qualitative} shows the qualitative results for scene graph generation (SGDET) of Visual Genome dataset.
Although some other proposals are also meaningful, we only demonstrate 9 relationships with the highest confidence scores and the generated scene graph due to space limitations in Fig.~\ref{fig:vg_qualitative}. 
Blue boxes are the subject boxes while orange boxes are the object boxes. 
Attention scores are displayed in the same color as boxes.
The overlap of subject and object attention is shown in white. 
The ground truth annotations of the two images are demonstrated in Fig.~\ref{fig:vg_gt}.
For brevity, we only show the bounding boxes of the entities that appear in the annotated triplets.

For the first image (with the car and building), we can assume that the 9 output triplets are all correct. 
The prediction \texttt{$<$car-in front of-building$>$} indicates that RelTR can understand spatial relationships from 2D image to some extent (\texttt{in front of} is not a high-frequent predicate in Visual Genome). 
However, R$@9$ of the first image is only $5/12=41.7$ because of the preferences in the ground truth triplet annotations. 
This phenomenon is more evident in the second image (with the woman and computer). 
Note that in the used Visual Genome-150 split~\cite{xu2017scene} there is no \texttt{computer} class but only \texttt{laptop} class.
6 out of 9 predictions from RelTR can be considered valid whereas R$@9$ is 0 due to the labeling preference. 
Sometimes RelTR outputs some duplicate triplets such as \texttt{$<$woman-wearing-jean$>$} and \texttt{$<$woman-looking at-laptop$>$} in the second image.
Along with the output results, RelTR also shows the regions of interest for the output relationships, making the behavior of the model easier to interpret.

%%%%%%%%%%%%%%%%%%

\begin{figure*}[http]
\centering
\includegraphics[width=0.84\linewidth]{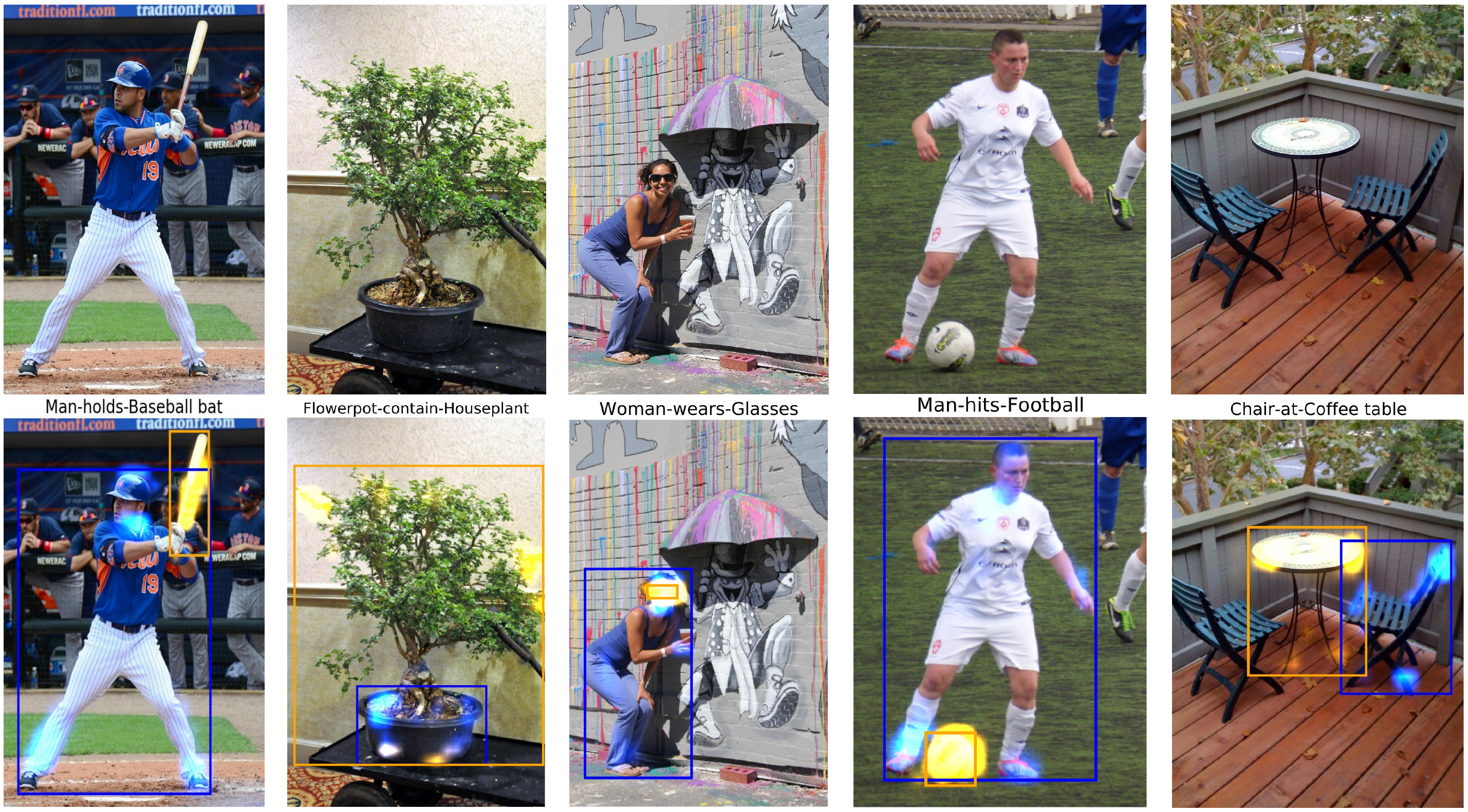}
\caption{Qualitative results for scene graph generation of Open Images V6. Different from the dense triplets in the annotations of VG, each image from Open Images V6 is labeled with 2.8 triplets on average. Although Open Images V6 contains more entity classes, the image scenarios are simpler compared to Visual Genome. Therefore, only the top-1 triplets are shown in the second row while the original images are in the first row. Boxes and attention scores of subjects are also colored with blue while objects with orange. RelTR demonstrates the excellent quality of its confident triplet proposals. }
\label{fig:oi6}
\end{figure*}

The qualitative results of SGDET for Open Images V6 are shown in Fig.~\ref{fig:oi6}. Different from the dense triplets in the annotations of VG, each image from Open Images V6 is labeled with 2.8 triplets on average. Therefore, we only show the most confident triplet from predictions for each image.

\section{Conclusion}
\label{sec:con}

In this paper, based on Transformer's encoder-decoder architecture, we propose a novel one-stage end-to-end framework for scene graph generation, RelTR. 
Given a fixed number of coupled subject and object queries, a fixed-size set of relationships is directly inferred based on visual appearance only, using different attention mechanisms in the triplet decoder of RelTR.
An IoU-based assignment strategy is proposed to optimize the triplet prediction-ground truth assignment during the model training. 
Compared with other state-of-the-art scene graph generation methods, RelTR achieves state-of-the-art performance on three datasets of different scales, with balanced performance on different evaluation metrics.
In contrast to  previous two-stage models, our approach does not require labeling predicates between all possible subject-object pairs but rather captures the triplets of interest through  attention mechanisms.
This results in the efficient and rapid inference of RelTR.
Moreover, our visual-based RelTR is easy to implement and  has the potential to be extended to an unbiased scene graph generation approach by using prior information.

% % use section* for acknowledgment
\ifCLASSOPTIONcompsoc
  \section*{Acknowledgments}
\else
  \section*{Acknowledgment}
\fi
This work has been supported by the Federal Ministry
of Education and Research (BMBF), under the
project LeibnizKILabor (grant no. 01DD20003), the Center for Digital Innovations (ZDIN) and the Deutsche Forschungsgemeinschaft (DFG) under Germany’s Excellence Strategy within the Cluster of Excellence PhoenixD (EXC 2122).

% % Can use something like this to put references on a page
% % by themselves when using endfloat and the captionsoff option.
% \ifCLASSOPTIONcaptionsoff
%   \newpage
% \fi

% \clearpage
 {
 \bibliographystyle{IEEEtran}
 \bibliography{egbib}

\begin{thebibliography}{10}
\providecommand{\url}[1]{#1}
\csname url@samestyle\endcsname
\providecommand{\newblock}{\relax}
\providecommand{\bibinfo}[2]{#2}
\providecommand{\BIBentrySTDinterwordspacing}{\spaceskip=0pt\relax}
\providecommand{\BIBentryALTinterwordstretchfactor}{4}
\providecommand{\BIBentryALTinterwordspacing}{\spaceskip=\fontdimen2\font plus
\BIBentryALTinterwordstretchfactor\fontdimen3\font minus
  \fontdimen4\font\relax}
\providecommand{\BIBforeignlanguage}[2]{{%
\expandafter\ifx\csname l@#1\endcsname\relax
\typeout{** WARNING: IEEEtran.bst: No hyphenation pattern has been}%
\typeout{** loaded for the language `#1'. Using the pattern for}%
\typeout{** the default language instead.}%
\else
\language=\csname l@#1\endcsname
\fi
#2}}
\providecommand{\BIBdecl}{\relax}
\BIBdecl

\bibitem{johnson2015image}
J.~Johnson, R.~Krishna, M.~Stark, L.-J. Li, D.~Shamma, M.~Bernstein, and
  L.~Fei-Fei, ``Image retrieval using scene graphs,'' in \emph{Proceedings of
  the IEEE conference on computer vision and pattern recognition}, 2015, pp.
  3668--3678.

\bibitem{lu2016visual}
C.~Lu, R.~Krishna, M.~Bernstein, and L.~Fei-Fei, ``Visual relationship
  detection with language priors,'' in \emph{European conference on computer
  vision}.\hskip 1em plus 0.5em minus 0.4em\relax Springer, 2016, pp. 852--869.

\bibitem{Nguyen_2021_ICCV}
K.~Nguyen, S.~Tripathi, B.~Du, T.~Guha, and T.~Q. Nguyen, ``In defense of scene
  graphs for image captioning,'' in \emph{Proceedings of the IEEE/CVF
  International Conference on Computer Vision (ICCV)}, 2021, pp. 1407--1416.

\bibitem{gao2018image}
L.~Gao, B.~Wang, and W.~Wang, ``Image captioning with scene-graph based
  semantic concepts,'' in \emph{Proceedings of the 2018 10th International
  Conference on Machine Learning and Computing}, 2018, pp. 225--229.

\bibitem{johnson2017inferring}
J.~Johnson, B.~Hariharan, L.~Van Der~Maaten, J.~Hoffman, L.~Fei-Fei,
  C.~Lawrence~Zitnick, and R.~Girshick, ``Inferring and executing programs for
  visual reasoning,'' in \emph{Proceedings of the IEEE International Conference
  on Computer Vision}, 2017, pp. 2989--2998.

\bibitem{ashual2019specifying}
O.~Ashual and L.~Wolf, ``Specifying object attributes and relations in
  interactive scene generation,'' in \emph{Proceedings of the IEEE/CVF
  International Conference on Computer Vision}, 2019, pp. 4561--4569.

\bibitem{johnson2018image}
J.~Johnson, A.~Gupta, and L.~Fei-Fei, ``Image generation from scene graphs,''
  in \emph{Proceedings of the IEEE conference on computer vision and pattern
  recognition}, 2018, pp. 1219--1228.

\bibitem{ren2016faster}
S.~Ren, K.~He, R.~Girshick, and J.~Sun, ``Faster r-cnn: towards real-time
  object detection with region proposal networks,'' \emph{IEEE transactions on
  pattern analysis and machine intelligence}, vol.~39, no.~6, pp. 1137--1149,
  2016.

\bibitem{zellers2018neural}
R.~Zellers, M.~Yatskar, S.~Thomson, and Y.~Choi, ``Neural motifs: Scene graph
  parsing with global context,'' in \emph{Proceedings of the IEEE Conference on
  Computer Vision and Pattern Recognition}, 2018, pp. 5831--5840.

\bibitem{chen2019knowledge}
T.~Chen, W.~Yu, R.~Chen, and L.~Lin, ``Knowledge-embedded routing network for
  scene graph generation,'' in \emph{Proceedings of the IEEE/CVF Conference on
  Computer Vision and Pattern Recognition}, 2019, pp. 6163--6171.

\bibitem{yu2017visual}
R.~Yu, A.~Li, V.~I. Morariu, and L.~S. Davis, ``Visual relationship detection
  with internal and external linguistic knowledge distillation,'' in
  \emph{Proceedings of the IEEE international conference on computer vision},
  2017, pp. 1974--1982.

\bibitem{zareian2020bridging}
A.~Zareian, S.~Karaman, and S.-F. Chang, ``Bridging knowledge graphs to
  generate scene graphs,'' in \emph{European Conference on Computer Vision},
  2020, pp. 606--623.

\bibitem{gu2019scene}
J.~Gu, H.~Zhao, Z.~Lin, S.~Li, J.~Cai, and M.~Ling, ``Scene graph generation
  with external knowledge and image reconstruction,'' in \emph{Proceedings of
  the IEEE/CVF Conference on Computer Vision and Pattern Recognition}, 2019,
  pp. 1969--1978.

\bibitem{law2018cornernet}
H.~Law and J.~Deng, ``Cornernet: Detecting objects as paired keypoints,'' in
  \emph{Proceedings of the European conference on computer vision (ECCV)},
  2018, pp. 734--750.

\bibitem{tian2019fcos}
Z.~Tian, C.~Shen, H.~Chen, and T.~He, ``Fcos: Fully convolutional one-stage
  object detection,'' in \emph{Proceedings of the IEEE/CVF international
  conference on computer vision}, 2019, pp. 9627--9636.

\bibitem{zhou2019objects}
X.~Zhou, D.~Wang, and P.~Kr{\"a}henb{\"u}hl, ``Objects as points,'' \emph{arXiv
  preprint arXiv:1904.07850}, 2019.

\bibitem{sun2021makes}
P.~Sun, Y.~Jiang, E.~Xie, W.~Shao, Z.~Yuan, C.~Wang, and P.~Luo, ``What makes
  for end-to-end object detection?'' in \emph{International Conference on
  Machine Learning}.\hskip 1em plus 0.5em minus 0.4em\relax PMLR, 2021, pp.
  9934--9944.

\bibitem{carion2020end}
N.~Carion, F.~Massa, G.~Synnaeve, N.~Usunier, A.~Kirillov, and S.~Zagoruyko,
  ``End-to-end object detection with transformers,'' in \emph{European
  Conference on Computer Vision}.\hskip 1em plus 0.5em minus 0.4em\relax
  Springer, 2020, pp. 213--229.

\bibitem{krishna2017visual}
R.~Krishna, Y.~Zhu, O.~Groth, J.~Johnson, K.~Hata, J.~Kravitz, S.~Chen,
  Y.~Kalantidis, L.-J. Li, D.~A. Shamma \emph{et~al.}, ``Visual genome:
  Connecting language and vision using crowdsourced dense image annotations,''
  \emph{International journal of computer vision}, vol. 123, no.~1, pp. 32--73,
  2017.

\bibitem{kuznetsova2020open}
A.~Kuznetsova, H.~Rom, N.~Alldrin, J.~Uijlings, I.~Krasin, J.~Pont-Tuset,
  S.~Kamali, S.~Popov, M.~Malloci, A.~Kolesnikov \emph{et~al.}, ``The open
  images dataset v4,'' \emph{International Journal of Computer Vision}, vol.
  128, no.~7, pp. 1956--1981, 2020.

\bibitem{yang2019auto}
X.~Yang, K.~Tang, H.~Zhang, and J.~Cai, ``Auto-encoding scene graphs for image
  captioning,'' in \emph{Proceedings of the IEEE/CVF Conference on Computer
  Vision and Pattern Recognition}, 2019, pp. 10\,685--10\,694.

\bibitem{gu2019unpaired}
J.~Gu, S.~Joty, J.~Cai, H.~Zhao, X.~Yang, and G.~Wang, ``Unpaired image
  captioning via scene graph alignments,'' in \emph{Proceedings of the IEEE/CVF
  International Conference on Computer Vision}, 2019, pp. 10\,323--10\,332.

\bibitem{lee2019learning}
K.-H. Lee, H.~Palangi, X.~Chen, H.~Hu, and J.~Gao, ``Learning visual relation
  priors for image-text matching and image captioning with neural scene graph
  generators,'' \emph{arXiv preprint arXiv:1909.09953}, 2019.

\bibitem{shi2019explainable}
J.~Shi, H.~Zhang, and J.~Li, ``Explainable and explicit visual reasoning over
  scene graphs,'' in \emph{Proceedings of the IEEE/CVF Conference on Computer
  Vision and Pattern Recognition}, 2019, pp. 8376--8384.

\bibitem{lee2019visual}
S.~Lee, J.-W. Kim, Y.~Oh, and J.~H. Jeon, ``Visual question answering over
  scene graph,'' in \emph{International Conference on Graph Computing (GC)},
  2019, pp. 45--50.

\bibitem{li2019pastegan}
Y.~Li, T.~Ma, Y.~Bai, N.~Duan, S.~Wei, and X.~Wang, ``Pastegan: A
  semi-parametric method to generate image from scene graph,'' \emph{Advances
  in Neural Information Processing Systems}, vol.~32, pp. 3948--3958, 2019.

\bibitem{talavera2019layout}
A.~Talavera, D.~S. Tan, A.~Azcarraga, and K.-L. Hua, ``Layout and context
  understanding for image synthesis with scene graphs,'' in \emph{IEEE
  International Conference on Image Processing (ICIP)}, 2019, pp. 1905--1909.

\bibitem{galleguillos2008object}
C.~Galleguillos, A.~Rabinovich, and S.~Belongie, ``Object categorization using
  co-occurrence, location and appearance,'' in \emph{2008 IEEE Conference on
  Computer Vision and Pattern Recognition}.\hskip 1em plus 0.5em minus
  0.4em\relax IEEE, 2008, pp. 1--8.

\bibitem{gould2008multi}
S.~Gould, J.~Rodgers, D.~Cohen, G.~Elidan, and D.~Koller, ``Multi-class
  segmentation with relative location prior,'' \emph{International journal of
  computer vision}, vol.~80, no.~3, pp. 300--316, 2008.

\bibitem{cong2020nodis}
Y.~Cong, H.~Ackermann, W.~Liao, M.~Y. Yang, and B.~Rosenhahn, ``Nodis: Neural
  ordinary differential scene understanding,'' in \emph{Proceedings of the
  European Conference on Computer Vision (ECCV)}, 2020, pp. 636--653.

\bibitem{wang2019exploring}
W.~Wang, R.~Wang, S.~Shan, and X.~Chen, ``Exploring context and visual pattern
  of relationship for scene graph generation,'' in \emph{Proceedings of the
  IEEE/CVF Conference on Computer Vision and Pattern Recognition}, 2019, pp.
  8188--8197.

\bibitem{shi2021simple}
J.~Shi, Y.~Zhong, N.~Xu, Y.~Li, and C.~Xu, ``A simple baseline for
  weakly-supervised scene graph generation,'' in \emph{Proceedings of the
  IEEE/CVF International Conference on Computer Vision}, 2021, pp.
  16\,393--16\,402.

\bibitem{wang2021topic}
W.~Wang, R.~Wang, and X.~Chen, ``Topic scene graph generation by attention
  distillation from caption,'' in \emph{Proceedings of the IEEE/CVF
  International Conference on Computer Vision}, 2021, pp. 15\,900--15\,910.

\bibitem{lu2021context}
Y.~Lu, H.~Rai, J.~Chang, B.~Knyazev, G.~Yu, S.~Shekhar, G.~W. Taylor, and
  M.~Volkovs, ``Context-aware scene graph generation with seq2seq
  transformers,'' in \emph{Proceedings of the IEEE/CVF International Conference
  on Computer Vision}, 2021, pp. 15\,931--15\,941.

\bibitem{tang2019learning}
K.~Tang, H.~Zhang, B.~Wu, W.~Luo, and W.~Liu, ``Learning to compose dynamic
  tree structures for visual contexts,'' in \emph{Proceedings of the IEEE/CVF
  Conference on Computer Vision and Pattern Recognition}, 2019, pp. 6619--6628.

\bibitem{chen2019counterfactual}
L.~Chen, H.~Zhang, J.~Xiao, X.~He, S.~Pu, and S.-F. Chang, ``Counterfactual
  critic multi-agent training for scene graph generation,'' in
  \emph{Proceedings of the IEEE/CVF International Conference on Computer
  Vision}, 2019, pp. 4613--4623.

\bibitem{chiou2021recovering}
M.-J. Chiou, H.~Ding, H.~Yan, C.~Wang, R.~Zimmermann, and J.~Feng, ``Recovering
  the unbiased scene graphs from the biased ones,'' in \emph{Proceedings of the
  29th ACM International Conference on Multimedia}, 2021, pp. 1581--1590.

\bibitem{ji2020action}
J.~Ji, R.~Krishna, L.~Fei-Fei, and J.~C. Niebles, ``Action genome: Actions as
  compositions of spatio-temporal scene graphs,'' in \emph{Proceedings of the
  IEEE/CVF Conference on Computer Vision and Pattern Recognition}, 2020, pp.
  10\,236--10\,247.

\bibitem{cong2021spatial}
Y.~Cong, W.~Liao, H.~Ackermann, B.~Rosenhahn, and M.~Y. Yang,
  ``Spatial-temporal transformer for dynamic scene graph generation,'' in
  \emph{Proceedings of the IEEE/CVF International Conference on Computer
  Vision}, 2021, pp. 16\,372--16\,382.

\bibitem{teng2021target}
Y.~Teng, L.~Wang, Z.~Li, and G.~Wu, ``Target adaptive context aggregation for
  video scene graph generation,'' in \emph{Proceedings of the IEEE/CVF
  International Conference on Computer Vision}, 2021, pp. 13\,688--13\,697.

\bibitem{lu2021multi}
Y.~Lu, C.~Chang, H.~Rai, G.~Yu, and M.~Volkovs, ``Multi-view scene graph
  generation in videos,'' in \emph{International Challenge on Activity
  Recognition (ActivityNet) CVPR 2021 Workshop}, vol.~3, 2021.

\bibitem{suhail2021energy}
M.~Suhail, A.~Mittal, B.~Siddiquie, C.~Broaddus, J.~Eledath, G.~Medioni, and
  L.~Sigal, ``Energy-based learning for scene graph generation,'' in
  \emph{Proceedings of the IEEE/CVF Conference on Computer Vision and Pattern
  Recognition}, 2021, pp. 13\,936--13\,945.

\bibitem{yan2020pcpl}
S.~Yan, C.~Shen, Z.~Jin, J.~Huang, R.~Jiang, Y.~Chen, and X.-S. Hua, ``Pcpl:
  Predicate-correlation perception learning for unbiased scene graph
  generation,'' in \emph{Proceedings of the 28th ACM International Conference
  on Multimedia}, 2020, pp. 265--273.

\bibitem{guo2021general}
Y.~Guo, L.~Gao, X.~Wang, Y.~Hu, X.~Xu, X.~Lu, H.~T. Shen, and J.~Song, ``From
  general to specific: Informative scene graph generation via balance
  adjustment,'' in \emph{Proceedings of the IEEE/CVF International Conference
  on Computer Vision}, 2021, pp. 16\,383--16\,392.

\bibitem{desai2021learning}
A.~Desai, T.-Y. Wu, S.~Tripathi, and N.~Vasconcelos, ``Learning of visual
  relations: The devil is in the tails,'' in \emph{Proceedings of the IEEE/CVF
  International Conference on Computer Vision}, 2021, pp. 15\,404--15\,413.

\bibitem{xu2017scene}
D.~Xu, Y.~Zhu, C.~B. Choy, and L.~Fei-Fei, ``Scene graph generation by
  iterative message passing,'' in \emph{Proceedings of the IEEE conference on
  computer vision and pattern recognition}, 2017, pp. 5410--5419.

\bibitem{li2017scene}
Y.~Li, W.~Ouyang, B.~Zhou, K.~Wang, and X.~Wang, ``Scene graph generation from
  objects, phrases and region captions,'' in \emph{Proceedings of the IEEE
  international conference on computer vision}, 2017, pp. 1261--1270.

\bibitem{yang2018graph}
J.~Yang, J.~Lu, S.~Lee, D.~Batra, and D.~Parikh, ``Graph r-cnn for scene graph
  generation,'' in \emph{Proceedings of the European conference on computer
  vision (ECCV)}, 2018, pp. 670--685.

\bibitem{li2018factorizable}
Y.~Li, W.~Ouyang, B.~Zhou, J.~Shi, C.~Zhang, and X.~Wang, ``Factorizable net:
  an efficient subgraph-based framework for scene graph generation,'' in
  \emph{Proceedings of the European Conference on Computer Vision (ECCV)},
  2018, pp. 335--351.

\bibitem{li2021bipartite}
R.~Li, S.~Zhang, B.~Wan, and X.~He, ``Bipartite graph network with adaptive
  message passing for unbiased scene graph generation,'' in \emph{Proceedings
  of the IEEE/CVF Conference on Computer Vision and Pattern Recognition}, 2021,
  pp. 11\,109--11\,119.

\bibitem{lin2020gps}
X.~Lin, C.~Ding, J.~Zeng, and D.~Tao, ``Gps-net: Graph property sensing network
  for scene graph generation,'' in \emph{Proceedings of the IEEE/CVF Conference
  on Computer Vision and Pattern Recognition}, 2020, pp. 3746--3753.

\bibitem{herzig2018mapping}
R.~Herzig, M.~Raboh, G.~Chechik, J.~Berant, and A.~Globerson, ``Mapping images
  to scene graphs with permutation-invariant structured prediction,''
  \emph{Advances in Neural Information Processing Systems}, vol.~31, pp.
  7211--7221, 2018.

\bibitem{qi2019attentive}
M.~Qi, W.~Li, Z.~Yang, Y.~Wang, and J.~Luo, ``Attentive relational networks for
  mapping images to scene graphs,'' in \emph{Proceedings of the IEEE/CVF
  Conference on Computer Vision and Pattern Recognition}, 2019, pp. 3957--3966.

\bibitem{vaswani2017attention}
A.~Vaswani, N.~Shazeer, N.~Parmar, J.~Uszkoreit, L.~Jones, A.~N. Gomez,
  {\L}.~Kaiser, and I.~Polosukhin, ``Attention is all you need,'' in
  \emph{Advances in neural information processing systems}, 2017, pp.
  5998--6008.

\bibitem{dhingra2021bgt}
N.~Dhingra, F.~Ritter, and A.~Kunz, ``Bgt-net: Bidirectional gru transformer
  network for scene graph generation,'' in \emph{Proceedings of the IEEE/CVF
  Conference on Computer Vision and Pattern Recognition}, 2021, pp. 2150--2159.

\bibitem{koner2020relation}
R.~Koner, P.~Sinhamahapatra, and V.~Tresp, ``Relation transformer network,''
  \emph{arXiv preprint arXiv:2004.06193}, 2020.

\bibitem{chen2022reltransformer}
J.~Chen, A.~Agarwal, S.~Abdelkarim, D.~Zhu, and M.~Elhoseiny, ``Reltransformer:
  A transformer-based long-tail visual relationship recognition,'' in
  \emph{Proceedings of the IEEE/CVF Conference on Computer Vision and Pattern
  Recognition}, 2022, pp. 19\,507--19\,517.

\bibitem{gkanatsios2019attention}
N.~Gkanatsios, V.~Pitsikalis, P.~Koutras, and P.~Maragos,
  ``Attention-translation-relation network for scalable scene graph
  generation,'' in \emph{Proceedings of the IEEE/CVF International Conference
  on Computer Vision Workshops}, 2019, pp. 0--0.

\bibitem{cui2018context}
Z.~Cui, C.~Xu, W.~Zheng, and J.~Yang, ``Context-dependent diffusion network for
  visual relationship detection,'' in \emph{Proceedings of the 26th ACM
  international conference on Multimedia}, 2018, pp. 1475--1482.

\bibitem{yu2020cogtree}
J.~Yu, Y.~Chai, Y.~Wang, Y.~Hu, and Q.~Wu, ``Cogtree: Cognition tree loss for
  unbiased scene graph generation,'' \emph{arXiv preprint arXiv:2009.07526},
  2020.

\bibitem{zhang2019graphical}
J.~Zhang, K.~J. Shih, A.~Elgammal, A.~Tao, and B.~Catanzaro, ``Graphical
  contrastive losses for scene graph parsing,'' in \emph{Proceedings of the
  IEEE/CVF Conference on Computer Vision and Pattern Recognition}, 2019, pp.
  11\,535--11\,543.

\bibitem{dai2017detecting}
B.~Dai, Y.~Zhang, and D.~Lin, ``Detecting visual relationships with deep
  relational networks,'' in \emph{Proceedings of the IEEE conference on
  computer vision and Pattern recognition}, 2017, pp. 3076--3086.

\bibitem{liu2021fully}
H.~Liu, N.~Yan, M.~Mortazavi, and B.~Bhanu, ``Fully convolutional scene graph
  generation,'' in \emph{Proceedings of the IEEE/CVF Conference on Computer
  Vision and Pattern Recognition}, 2021, pp. 11\,546--11\,556.

\bibitem{li2022sgtr}
R.~Li, S.~Zhang, and X.~He, ``Sgtr: End-to-end scene graph generation with
  transformer,'' in \emph{Proceedings of the IEEE/CVF Conference on Computer
  Vision and Pattern Recognition}, 2022, pp. 19\,486--19\,496.

\bibitem{yuan2022rlip}
H.~Yuan, J.~Jiang, S.~Albanie, T.~Feng, Z.~Huang, D.~Ni, and M.~Tang, ``Rlip:
  Relational language-image pre-training for human-object interaction
  detection,'' \emph{arXiv preprint arXiv:2209.01814}, 2022.

\bibitem{kim2021hotr}
B.~Kim, J.~Lee, J.~Kang, E.-S. Kim, and H.~J. Kim, ``Hotr: End-to-end
  human-object interaction detection with transformers,'' in \emph{Proceedings
  of the IEEE/CVF Conference on Computer Vision and Pattern Recognition}, 2021,
  pp. 74--83.

\bibitem{wang2021end}
Y.~Wang, Z.~Xu, X.~Wang, C.~Shen, B.~Cheng, H.~Shen, and H.~Xia, ``End-to-end
  video instance segmentation with transformers,'' in \emph{Proceedings of the
  IEEE/CVF Conference on Computer Vision and Pattern Recognition}, 2021, pp.
  8741--8750.

\bibitem{liu2021cptr}
W.~Liu, S.~Chen, L.~Guo, X.~Zhu, and J.~Liu, ``Cptr: Full transformer network
  for image captioning,'' \emph{arXiv preprint arXiv:2101.10804}, 2021.

\bibitem{zeng2021motr}
F.~Zeng, B.~Dong, T.~Wang, C.~Chen, X.~Zhang, and Y.~Wei, ``Motr: End-to-end
  multiple-object tracking with transformer,'' \emph{arXiv preprint
  arXiv:2105.03247}, 2021.

\bibitem{zhu2020deformable}
X.~Zhu, W.~Su, L.~Lu, B.~Li, X.~Wang, and J.~Dai, ``Deformable detr: Deformable
  transformers for end-to-end object detection,'' in \emph{International
  Conference on Learning Representations (ICLR)}, 2021.

\bibitem{yao2021efficient}
Z.~Yao, J.~Ai, B.~Li, and C.~Zhang, ``Efficient detr: Improving end-to-end
  object detector with dense prior,'' \emph{arXiv preprint arXiv:2104.01318},
  2021.

\bibitem{zou2021end}
C.~Zou, B.~Wang, Y.~Hu, J.~Liu, Q.~Wu, Y.~Zhao, B.~Li, C.~Zhang, C.~Zhang,
  Y.~Wei \emph{et~al.}, ``End-to-end human object interaction detection with
  hoi transformer,'' in \emph{Proceedings of the IEEE/CVF Conference on
  Computer Vision and Pattern Recognition}, 2021, pp. 11\,825--11\,834.

\bibitem{stewart2016end}
R.~Stewart, M.~Andriluka, and A.~Y. Ng, ``End-to-end people detection in
  crowded scenes,'' in \emph{Proceedings of the IEEE conference on computer
  vision and pattern recognition}, 2016, pp. 2325--2333.

\bibitem{rezatofighi2019generalized}
H.~Rezatofighi, N.~Tsoi, J.~Gwak, A.~Sadeghian, I.~Reid, and S.~Savarese,
  ``Generalized intersection over union: A metric and a loss for bounding box
  regression,'' in \emph{Proceedings of the IEEE/CVF Conference on Computer
  Vision and Pattern Recognition}, 2019, pp. 658--666.

\bibitem{tang2020unbiased}
K.~Tang, Y.~Niu, J.~Huang, J.~Shi, and H.~Zhang, ``Unbiased scene graph
  generation from biased training,'' in \emph{Proceedings of the IEEE/CVF
  Conference on Computer Vision and Pattern Recognition}, 2020, pp. 3716--3725.

\bibitem{loshchilov2017decoupled}
I.~Loshchilov and F.~Hutter, ``Decoupled weight decay regularization,'' in
  \emph{International Conference on Learning Representations (ICLR)}, 2019.

\bibitem{deng2009imagenet}
J.~Deng, W.~Dong, R.~Socher, L.-J. Li, K.~Li, and L.~Fei-Fei, ``Imagenet: A
  large-scale hierarchical image database,'' in \emph{2009 IEEE conference on
  computer vision and pattern recognition}.\hskip 1em plus 0.5em minus
  0.4em\relax Ieee, 2009, pp. 248--255.

\bibitem{lin2014microsoft}
T.-Y. Lin, M.~Maire, S.~Belongie, J.~Hays, P.~Perona, D.~Ramanan,
  P.~Doll{\'a}r, and C.~L. Zitnick, ``Microsoft coco: Common objects in
  context,'' in \emph{European conference on computer vision}.\hskip 1em plus
  0.5em minus 0.4em\relax Springer, 2014, pp. 740--755.

\bibitem{al2019character}
R.~Al-Rfou, D.~Choe, N.~Constant, M.~Guo, and L.~Jones, ``Character-level
  language modeling with deeper self-attention,'' in \emph{Proceedings of the
  AAAI Conference on Artificial Intelligence}, vol.~33, no.~01, 2019, pp.
  3159--3166.

\bibitem{zhang2017visual}
H.~Zhang, Z.~Kyaw, S.-F. Chang, and T.-S. Chua, ``Visual translation embedding
  network for visual relation detection,'' in \emph{Proceedings of the IEEE
  conference on computer vision and pattern recognition}, 2017, pp. 5532--5540.

\bibitem{newell2017pixels}
A.~Newell and J.~Deng, ``Pixels to graphs by associative embedding,''
  \emph{Advances in neural information processing systems}, vol.~30, 2017.

\bibitem{li2017vip}
Y.~Li, W.~Quyang, and X.~Wang, ``Vip-cnn: A visual phrase reasoning
  convolutional neural network for visual relationsip detection,'' \emph{arXiv
  preprint arXiv:1702.07191}, 2017.

\bibitem{liang2017deep}
X.~Liang, L.~Lee, and E.~P. Xing, ``Deep variation-structured reinforcement
  learning for visual relationship and attribute detection,'' in
  \emph{Proceedings of the IEEE conference on computer vision and pattern
  recognition}, 2017, pp. 848--857.

\bibitem{zhan2019exploring}
Y.~Zhan, J.~Yu, T.~Yu, and D.~Tao, ``On exploring undetermined relationships
  for visual relationship detection,'' in \emph{Proceedings of the IEEE/CVF
  Conference on Computer Vision and Pattern Recognition}, 2019, pp. 5128--5137.

\bibitem{yin2018zoom}
G.~Yin, L.~Sheng, B.~Liu, N.~Yu, X.~Wang, J.~Shao, and C.~C. Loy, ``Zoom-net:
  Mining deep feature interactions for visual relationship recognition,'' in
  \emph{Proceedings of the European Conference on Computer Vision (ECCV)},
  2018, pp. 322--338.

\bibitem{gupta2019lvis}
A.~Gupta, P.~Dollar, and R.~Girshick, ``Lvis: A dataset for large vocabulary
  instance segmentation,'' in \emph{Proceedings of the IEEE/CVF conference on
  computer vision and pattern recognition}, 2019, pp. 5356--5364.

\bibitem{menon2021longtail}
A.~K. Menon, S.~Jayasumana, A.~S. Rawat, H.~Jain, A.~Veit, and S.~Kumar,
  ``Long-tail learning via logit adjustment,'' in \emph{International
  Conference on Learning Representations}, 2021.

\bibitem{teng2022structured}
Y.~Teng and L.~Wang, ``Structured sparse r-cnn for direct scene graph
  generation,'' in \emph{Proceedings of the IEEE/CVF Conference on Computer
  Vision and Pattern Recognition}, 2022, pp. 19\,437--19\,446.

\end{thebibliography}
 }


\begin{IEEEbiography}[{\includegraphics[width=1in,height=1.25in,clip,keepaspectratio]{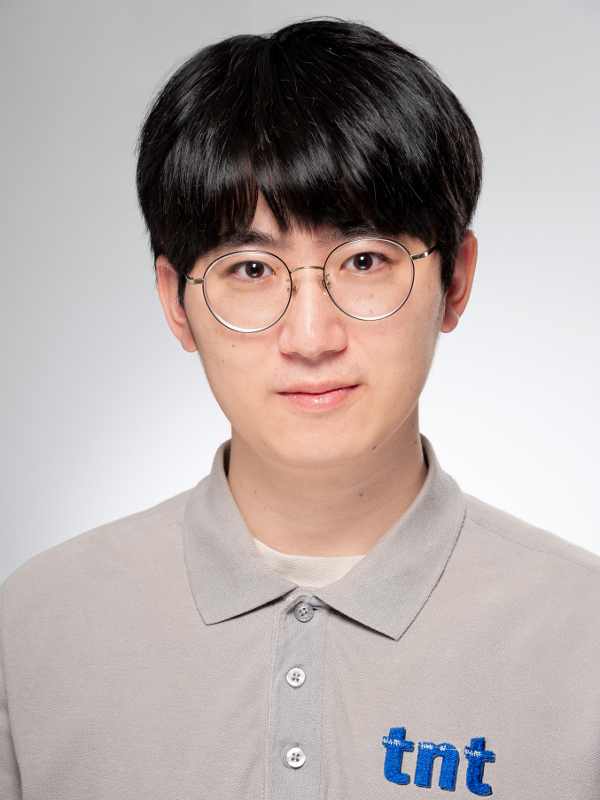}}]{Yuren Cong}
received his Bachelor degree at Hefei University of Technology in 2015. Then he studied Electrical Engineering and Information Technology at Leibniz University Hannover and received his Master degree in 2019. Since 2020 he has worked as a research assistant towards his Ph.D in the group of Prof. Rosenhahn. His research interests are in the fields of computer vision with specialization on scene graph generation.
	\end{IEEEbiography}
	
\begin{IEEEbiography}[{\includegraphics[width=1in,height=1.25in,clip,keepaspectratio]{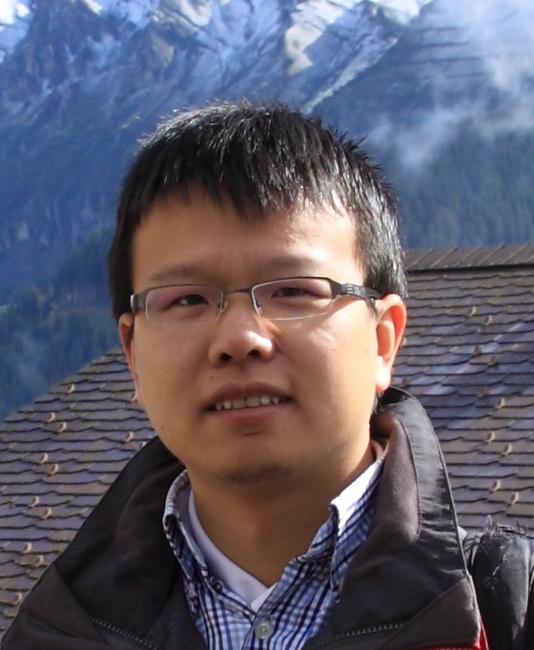}}]{Micheal Ying Yang}
	               is currently Assistant Professor
at University of Twente, The Netherlands, heading a group working
on scene understanding. He received the
PhD degree (summa cum laude) from University
of Bonn (Germany) in 2011. He received the
venia legendi in Computer Science from Leibniz
University Hannover in 2016. His research interests
are in the fields of computer vision and photogrammetry
with specialization on scene understanding and multi-modal learning. 
He has co-authored over 100 papers and organized several workshops in the last years.
He serves as Associate Editor of ISPRS
Journal of Photogrammetry and Remote Sensing, Co-chair of ISPRS
working group II/5 Temporal Geospatial Data Understanding, Program Chair of ISPRS Geospatial Week 2019, and recipient of ISPRS President’s Honorary Citation (2016), Best Science Paper Award at BMVC (2016), and The Willem Schermerhorn Award (2021).
He is regularly serving as program committee member of conferences and reviewer for international journals.
				\end{IEEEbiography}

\begin{IEEEbiography}[{\includegraphics[width=1in,height=1.25in,clip,keepaspectratio]{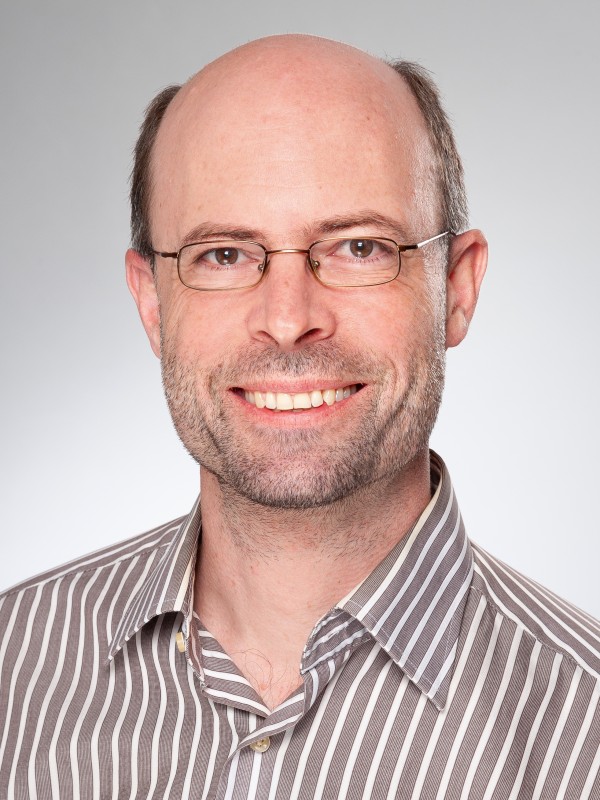}}]		{Bodo Rosenhahn} studied Computer Science (minor subject Medicine) at the University of
Kiel. He received the Dipl.-Inf. and Dr.-Ing. from the University of Kiel in 1999 and 2003,
respectively. From 10/2003 till 10/2005, he worked as PostDoc at the University of Auckland
(New Zealand), funded with a scholarship from the German Research Foundation (DFG). In
11/2005-08/2008 he worked as senior researcher at the Max-Planck Institute for Computer
Science. Since 09/2008 he is Full Professor at the Leibniz-University of
Hannover, heading a group on automated image interpretation. He has co-authored over
200 papers, holds 12 patents and organized several workshops and conferences in the last
years.
His works received several awards, including a DAGM-Prize 2002, the Dr.-Ing. Siegfried Werth Prize 2003, the DAGM-Main Prize 2005, the DAGM-Main Prize 2007, the Olympus-Prize 2007, and the Günter Enderle Award (Eurographics) 2017. 

	\end{IEEEbiography}

\end{document}